\documentclass{article} 
\usepackage{iclr2026_conference_custom,times}
\makeatletter
\ifdefined\@submissionfalse
    \@submissionfalse
\else
    \newif\if@submission
    \@submissionfalse
\fi
\makeatother 
\iclrfinalcopy
\usepackage{tikz}
\usetikzlibrary{matrix}
\usepackage{hyperref} 
\usepackage{amsmath,amssymb,amsfonts}
\usepackage{cleveref}
\usepackage{url}
\usepackage{multirow}
\usepackage{booktabs}
\usepackage{multicol}
\usepackage{graphicx}
\usepackage{svg}
\usepackage[most]{tcolorbox}
 \usepackage{subcaption} 
\usepackage{enumitem}
\usepackage{comment}
\usepackage{xcolor}
\definecolor{darkblue}{RGB}{0, 0, 139}   
\definecolor{lightblue}{RGB}{100, 149, 237} 
\definecolor{lightestblue}{RGB}{0, 0, 50}
\definecolor{boxgray}{RGB}{245, 245, 245}

\newtcolorbox{tokenforkbox}{
    colback=boxgray,
    colframe=gray!20,
    boxrule=0.5pt,
    arc=4pt,
    left=5pt,
    right=5pt,
    top=5pt,
    bottom=5pt,
    fontupper=\small\ttfamily,
    halign=left,        
    breakable,          
    lines before break=4, 
    enhanced            
}

\newtcbox{\tokLightBlueBox}{on line,
  colback=blue!15,
  colframe=blue!45,
  boxrule=0.4pt,
  arc=2pt,
  left=2pt,right=2pt,top=1pt,bottom=1pt,
  boxsep=0pt
}

\newtcbox{\tokDarkBlueBox}{on line,
  colback=yellow!20,
  colframe=yellow!85,
  boxrule=0.4pt,
  arc=2pt,
  left=2pt,right=2pt,top=1pt,bottom=1pt,
  boxsep=0pt
}

\newtcbox{\tokLightestBlueBox}{on line,
  colback=red!7,
  colframe=red!45,
  boxrule=0.4pt,
  arc=2pt,
left=2pt,right=2pt,top=1pt,bottom=1pt,
  boxsep=0pt
}

\usepackage{expl3}
\ExplSyntaxOn
\cs_new_eq:NN \tokenfork_old_textcolor \textcolor
\RenewDocumentCommand{\textcolor}{m m}{
  \str_case:nnF {#1}{
    {lightblue}{\tokLightBlueBox{#2}}
    {darkblue}{\tokDarkBlueBox{#2}}
    {lightestblue}{\tokLightestBlueBox{#2}}
  }{
    \tokenfork_old_textcolor{#1}{#2}
  }
}
\ExplSyntaxOff

\title{Multiplex Thinking: \\Reasoning via Token-wise Branch-and-Merge}

\author{
  Yao Tang$^{1}$\thanks{Work partially done during internship at Microsoft Research.}\quad\ \ Li Dong$^{2}$\quad\ \  Yaru Hao$^{2}$\quad\ \  Qingxiu Dong$^{2}$\quad\ \  Furu Wei$^{2}$\quad\ \  Jiatao Gu$^{1}$ \vspace{4pt}\\
  \vspace{2pt}
  $^{1}$University of Pennsylvania \qquad $^{2}$Microsoft Research \\
  \vspace{2pt}
  $^{1}$\texttt{\{tangyao,jgu32\}@seas.upenn.edu}
 \vspace{-10pt}
}

\NewDocumentCommand{\JG}{ mO{} }{\textcolor{teal}
{\textsuperscript{\textit{JG}}\textsf{\textbf{\small[#1]}}}}
\newcommand{\name}[1]{\textsc{LRL}}
\newcommand{\MODEL}[1]{\textsc{Multiplex Thinking}}
\newcommand{\Model}[1]{Multiplex Thinking}
\newcommand{\model}[1]{multiplex thinking}
\newcommand{\shortname}[1]{MPT}
\begin{document}

\newcommand{\tok}[2]{\textcolor{#1}{#2}\allowbreak}

\maketitle

\begin{abstract}
Large language models often solve complex reasoning tasks more effectively with Chain-of-Thought (CoT), but at the cost of long, low-bandwidth token sequences. Humans, by contrast, often reason softly by maintaining a distribution over plausible next steps. Motivated by this, we propose \textbf{\Model{}}, a stochastic soft reasoning mechanism that, at each thinking step, samples $K$ candidate tokens and aggregates their embeddings into a single continuous \emph{multiplex token}. This preserves the vocabulary embedding prior and the sampling dynamics of standard discrete generation, while inducing a tractable probability distribution over multiplex rollouts. Consequently, multiplex trajectories can be directly optimized with on-policy reinforcement learning (RL). 
Importantly, \model{} is self-adaptive: when the model is confident, the multiplex token is nearly discrete and behaves like standard CoT; when it is uncertain, it compactly represents multiple plausible next steps without increasing sequence length. Across challenging math reasoning benchmarks, \model{} consistently outperforms strong discrete CoT and RL baselines from Pass@1 through Pass@1024, while producing shorter sequences. The code and checkpoints are available at \href{https://github.com/GMLR-Penn/Multiplex-Thinking}{github.com/GMLR-Penn/Multiplex-Thinking}.
\end{abstract}

\vspace{5pt}
\section{Introduction}
Large Language Models (LLMs) have exhibited exceptional reasoning capabilities on a wide range of complex tasks, especially in mathematics and logical problem solving~\citep{cobbe2021training,lightman2023lets}. 
A simple and effective way to elicit such behavior is \emph{chain-of-thought} (CoT) prompting~\citep{wei2022chain}, which encourages the model to generate intermediate reasoning steps before producing the final answer. 
Beyond prompting, reinforcement learning (RL) can further improve reasoning by optimizing the model over diverse CoT rollouts using outcome- or process-level rewards~\citep{deepseekr1}, steering probability mass toward higher-reward reasoning trajectories.

\begin{figure}[tb]
    \centering
   \includegraphics[width=\linewidth,trim=160pt 125pt 230pt 140pt,clip]{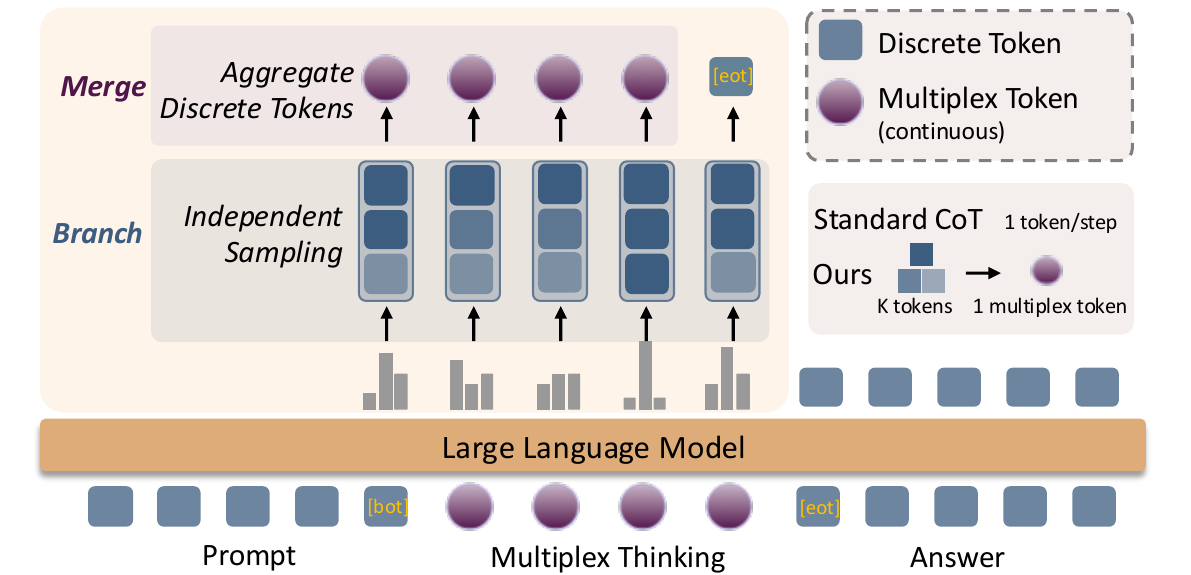}
    \caption{\textbf{Illustration of \Model{}}. The model first generates an initial probability distribution conditioned on the prompt and the begin of thinking token\texttt{[bot]}. Instead of committing to a single discrete token or a deterministic soft average, we conduct token-wise branching and merging by independently sampling $K$ discrete tokens and aggregate them into a continuous multiplex token. When one of the sampled discrete tokens is the end of thinking token \texttt{[eot]}, LLM continues conduct discrete decoding to give the answer. 
    The design of sampling-based continuous thinking bridges the gap between continuous representation and stochastic discrete sampling, allowing for effective on-policy exploration and further RL training. 
    }
    \label{fig:placeholder}
\end{figure}

However, both CoT prompting and RL on CoT rollouts are costly because they require generating long sequences of discrete tokens. Each rollout corresponds to a full, explicit reasoning trace, and exploring alternatives often resembles depth-first search (DFS): each sampled trace commits to a single trajectory before branching to others~\citep{zhu2025reasoningsuperpositiontheoreticalperspective}. This cost motivates \emph{continuous} reasoning tokens that can compactly encode a ``superposition'' over multiple candidate reasoning paths within a single token and decode in a more breadth-first search (BFS)-like manner~\citep{zhu2025reasoningsuperpositiontheoreticalperspective}. 

While existing continuous token approaches can reduce token cost, they are typically \emph{deterministic}: given the next-token logits, they map the distribution to a single continuous vector, e.g., a hidden-state token~\citep{hao2025coconut} or a probability-weighted embedding mixture~\citep{softthinking}.  Determinism collapses the token-level policy distribution, making decoded rollouts identical and thus limiting exploration. This characteristic is fundamentally misaligned with RL, where on-policy stochastic rollouts are crucial to enable LLMs to effectively learn from \emph{trial-and-error}s.  Therefore, we posit that continuous reasoning tokens should be \emph{stochastic and sampling-based}, retaining discrete sampling dynamics while operating in a compact continuous space.

To fill this gap, we propose \textbf{\Model{}}, a discrete-sampling-based continuous reasoning paradigm.
At each reasoning step, \model{} samples $K$ independent tokens over model’s token distribution, maps them to their vocabulary embeddings, and aggregates the embeddings into a single multiplex thinking token. When the logits are highly peaked (low entropy), the sampled tokens are likely to coincide, and the multiplex token effectively collapses to a standard single-token embedding. In contrast, when the logits have high entropy and exploration is desirable, the $K$ samples are more likely to differ, allowing the multiplex token to encode richer, multi-faceted information within one step. 
Crucially, \model{} preserves both the vocabulary embedding prior and the sampling behavior of discrete token generation, while introducing a continuous-like reasoning representation. Because the sampled tokens at each step are independently from the underlying token distribution, the probability of a multiplex token is the product of the probabilities of its sampled tokens. This factorization allows us to explicitly model the probability of an entire \model{} rollout and thereby optimize \model{} directly with RL.

Empirically, we show that \model{} consistently improves performance from Pass@1 to Pass@1024 across a range of challenging math reasoning benchmarks, surpassing strong discrete CoT and RL baselines. We further find that \model{} achieves higher accuracy while maintaining better token efficiency: responses are shorter on average, since a single multiplex token can encode richer information than a standard discrete token.

Our main contributions can be summarized as:
\begin{itemize}[leftmargin=10pt]
    \item We introduce \model{}, a token-efficient reasoning scheme that compresses multiple discrete CoT samples into continuous \emph{multiplex} tokens while retaining stochastic exploration and probabilistic semantics.
    \item We formalize \model{} as inducing a well-defined probability distribution over complete reasoning trajectories, enabling direct RL optimization over multiplex rollouts without paying the full token cost of long discrete CoT.
    \item We demonstrate consistent gains over strong discrete CoT and RL baselines across sampling budgets (Pass@1--Pass@1024), and provide analysis showing improved accuracy with shorter trajectories by compressing high-entropy reasoning steps.
\end{itemize}

\section{Background}

\paragraph{Chain-of-Thought (CoT) Reasoning} In standard CoT reasoning, LLMs are prompted to output intermediate thinking tokens before predicting the final answers to solve problems. Given a language model $\pi_\theta$ with a vocabulary $V$ and the embedding layer $E \in R^{|V| \times d}$, the $k$-th token in the vocabulary can be embedded as $E[k]$, also denoted as $e(k)$. With an input question $q=(q_1, q_2,\dots, q_L)$, language model $\pi_\theta$ first outputs the thinking sequence $t$ and then outputs the final answer $y$, where 
$
t_{i}\sim \pi_\theta(e(q),e(t_{<i}))
$ and $
y_{i}\sim \pi_\theta(e(q),e(t),e(y_{<i}))
$. While effective, standard CoT samples a discrete token at each reasoning step. This abandons the rich information of the distribution over the whole vocabulary, which further motivates research on reasoning over continous tokens to better preserve the information over the vocabulary.


\paragraph{Soft Thinking} Soft Thinking~\citep{softthinking} enhances LLM performance without fine-tuning by replacing discrete thinking tokens $t$ with continuous \emph{concept tokens} $c$. At the $i$-th thinking step, it constructs a concept token by using the model's next-token distribution $p_i=\pi_\theta(\cdot \mid e(q), c_{<i})$ as weights to aggregate token embeddings over the vocabulary $\mathcal{V}$:
$$c_i = \sum_{k \in V} p_i(k) e(k).$$


Although Soft Thinking effectively compresses reasoning information into continuous vectors, it is inherently \textit{deterministic}: given a context, the mapping from the logit distribution to $c_i$ is fixed. This lack of stochasticity prevents the model from exploring diverse reasoning paths, thereby limiting its potential to be optimized via reinforcement learning objectives that rely on trial-and-error.

\paragraph{Reinforcement Learning with Verifiable Rewards (RLVR)}
RLVR trains language models on tasks with verifiable answers with a reinforcement learning objective~\citep{tulu3}. Given a verifiable dataset consisting of question-answer pairs $\mathcal{D}=\{(q,y^\star)\}$ and an answer sampled from a language model $y \sim \pi_\theta( \cdot \mid q)$, a verifiable reward function $v$ can provide a reward $r=v(y,a)$ based on the ground-truth answer $y^\star$ and the sampled answer $y$. The model is trained to maximize the reward, given by:
$$
\mathcal{J}_\text{RLVR}=\mathbb{E}_{(q,y^\star)\sim\mathcal{D},y\sim \pi_\theta( \cdot \mid q)}[v(y,y^\star)].
$$

\section{\Model{}}

In this section, we propose \textbf{\model{}}, a reasoning paradigm effectively combining the information density of continuous representations with the probabilistic structure of discrete sampling. 
Based on multiplex thinking trajectories, we further introduce the reinforcement learning objective and provide entropy analysis of multiplex tokens. 

\subsection{Formulation}
Given a question $q$ sampled from dataset $\mathcal{D}$, language model $\pi_\theta$ samples a multiplex thinking trace $c=(c_1,c_2,\dots ,c_L)$ followed by a final answer $y$. Different from standard decoding where a discrete token is sampled at each step, \model{} constructs a mutliplex token with width $K$ by independently sampling $K$ times and aggregating the sampled discrete tokens.
Formally, at reasoning step $i$, we independently sample $K$ discrete tokens $k_{i,1}, k_{i,2}, \dots, k_{i,K}$ from the model's distribution $\pi_\theta(\cdot \mid e(q), c_{<i})$. 
We aggregate these discrete samples by averaging their one-hot vectors:
$$
{s}_i \;=\; \frac{1}{K}\sum_{j=1}^{K}{z}_{i,j},
$$
where $z_{i,j}$ is the one-hot vector corresponding to the discrete token $k_{i,j}$. When $K=1$, ${s}_i={z}_{i,1}$ collapses to a one-hot vector, and the multiplex token degenerates to a standard discrete token sampled from $\pi_\theta$. When $K\to\infty$, the empirical distribution of $K$ i.i.d.\ samples converges to the model's LM head distribution $\pi_\theta(\cdot\mid e(q),c_{<i})$. 

To obtain a continuous representation, we map ${s}_i$ through the embedding matrix $E \in \mathbb{R}^{V\times d}$, and we further define the continuous multiplex token by applying a vocabulary-space weighting $w_i \in \mathbb{R}^{V}$ to ${s}_i$:
$$
c_i
\;=\;
E^\top\!\left(
{s}_i \odot w_i
\right),
$$
where $\odot$ denotes element-wise multiplication. We consider two choices of $w_i$. \textbf{(i) Uniform averaging:} $w_i[v]=1$ for all vocabulary indices $v$, which recovers $c_i = E^\top s_i$ (the averaged embedding of sampled tokens). \textbf{(ii) LM-head reweighting:} we set 
$w_i[v] \;=\; K \cdot
\frac{\mathbf{1}\!\left[{s}_i[v] > 0\right]\cdot \pi_\theta\!\left(v \mid e(q), c_{<i}\right)}
{\sum_{u=1}^{V}\mathbf{1}\!\left[{s}_i[u] > 0\right]\cdot \pi_\theta\!\left(u \mid e(q), c_{<i}\right)},$ i.e., we only reweight tokens that appear in the sampled set and scale them according to the model’s LM-head probabilities.
Empirically, we find that uniform averaging over samples and LM-head reweighting lead to comparable performance in Section~\ref{sec:ablate_on_weight_or_unweight}. In our experiments, we adopt reweighting by default as it more directly reflects the model's confidence over the sampled candidates.

By independently sampling multiplex discrete tokens and aggregating these sampled tokens into a continuous representation, this design allows each $c_i$ to capture a stochastic ensemble of potential reasoning paths. When the distribution is sharp with low entropy, the samples collapse to the same token, reverting to standard discrete behavior. Conversely, high entropy distributions result in a diverse mixture, encoding exploration within a single continuous vector. 

The probability of generating a specific multiplex token factorizes due to the independence assumption. Consequently, the log-probability of the entire reasoning trace $c$ is the sum of the log-probabilities of all constituent discrete samples:
$$
 \log \pi(c |e( q)) = \sum_{i=1}^{|c|} \sum _{j=1}^K \log \pi_{\theta}(k_{i,j} | e(q), c_{<i}).
$$


\subsection{Reinforcement Learning Objective}

Leveraging the factorization above, we can directly optimize the model using Reinforcement Learning. We aim to maximize the expected reward of the generated answer $y$. The objective function is defined as:
$$
\mathcal{J}_{\text{RL}}(\theta) =\mathbb{E}_{\substack{(q,y^\star)\sim\mathcal{D}, \\ c\sim \pi_\theta(\cdot|q), \\ y\sim \pi_\theta(\cdot|q,c)}} \left[\left( \log \pi_{\theta}\left(c| e(q)\right) +  \log \pi_{\theta}\left(y | e(q), c\right)\right) \cdot v(y,y^\star)\right].
$$
This objective performs on-policy reinforcement learning over the joint generation process of the multiplex thinking trace $c$ and the final answer $y$,

 \subsection{Entropy of multiplex token}

To quantify the exploration capability of \model{}, we compare the entropy of the multiplex token against the standard discrete token. In standard Chain-of-Thought (CoT), the discrete token $t_i$ is sampled from the policy $\pi_\theta(\cdot \mid q, t_{<i})$ at the decoding step $i$. The entropy of this single-step sampling is given by the standard Shannon entropy:
$$
H_{\text{CoT}}(i) = -\sum_{v \in V} \pi_\theta(v \mid q, t_{<i}) \log \pi_\theta(v \mid q, t_{<i}).
$$
In contrast, a multiplex token is constructed by independently sampling $K$ tokens $\mathcal{K}_i=\{k_{i,1}, \dots, k_{i,K}\}$ from the distribution $\pi_\theta(\cdot \mid q, c_{<i})$. We conceptualize the generation of a multiplex token as a single \textit{integrated action} that selects a composite outcome $ (k_{i,1}, \dots, k_{i,K})$ from the augmented state space $|\mathcal{V}|^K$.
Under the assumption of sampling independence, we treat the multiplex token $c_i$ as a unified random variable and analyze the entropy of $c_i$ as a joint entropy, which is the sum of individual sampling entropies:

$$
H(\mathcal{K}_i) = K \cdot H(\pi_\theta(q,c_{<i})).
$$
We could observe that the entropy scales linearly with $K$, which corresponds to an exponential expansion of the effective exploration volume from $|\mathcal{V}|$ to $|\mathcal{V}|^K$.
While standard CoT commits to a single discrete path, \Model{} leverages the high-capacity continuous space to encode a `superposition' of $K$ paths simultaneously~\citep{zhu2025reasoningsuperpositiontheoreticalperspective}. This allows the model to defer discrete decisions and retain probabilistic diversity within the reasoning trace. This property is particularly advantageous for reinforcement learning, as it provides a richer training signal. We also provide empirical analysis in Experiments~\ref{sec:experiment:entropy}.

\section{Experiments}
In this section, we empirically evaluate \model{} against discrete reasoning approaches and competitive continuous reasoning baselines. Our evaluation focuses on both Pass@1 accuracy and test-time scaling (Pass@1--Pass@1024), measuring effectiveness at small budgets as well as the exploration gains from increased rollouts.
Overall, \model{} achieves the best Pass@1 performance in major settings and exhibits stronger Pass@k scaling than discrete baselines.

\subsection{Experimental Setups}

\paragraph{Implementation} We implement \model{} on top of two open-source reasoning backbones: \texttt{DeepSeek-R1-Distill-Qwen-1.5B} and \texttt{DeepSeek-R1-Distill-Qwen-7B}. The models are optimized using Group Relative Policy Optimization (GRPO)~\citep{grpo}.
We train for 300 steps with a global batch size of 128 questions, a learning rate of $1\times 10^{-6}$, and zero KL penalty and entropy penalty. A maximum response length of 4096 tokens is enforced for all training and evaluation stages.

During training, we generate 8 rollout samples per question with the temperature of 1.0 and the top-p of 1.0 to train LLMs with on-policy RL. 
During evaluation, we use a top-p of 0.95 to evaluate the Pass@1 performance, and these results are averaged on 64 runs. We also measure the Pass@$k$ performance~\citep{chen2021evaluatinglargelanguagemodels} for $k\in\{1,2,4,\dots,1024\}$ to fully investigate the exploration upper limit of different methods with the top-p of 1.0. Pass@$k$ measures the probability that at least one correct solution exists among $k$ sampled trajectories, serving as a proxy for the model's potential to discover a valid solution within a given exploration budget. The experiment evaluating the Pass@$k$ performance for $k\in\{1,2,4,\dots,1024\}$ aims to provide a thorough empirical study on how the exploration space of \model{} scales with exploration budget in comparison with discrete baselines. Results evaluating Pass@$k$ for $k\in\{1,2,4,\dots,1024\}$ are computed by boostrapping for 1,000 times on a total of 1,024 runs. More implementation details are provided in Appendix~\ref{Appendix:experiment_details}.

\paragraph{Datasets} 
The training set is DeepScaleR-Preview-Dataset~\citep{deepscaler2025} consisting of approximately 40,000 unique problem-answer pairs. For evaluation, we use six challenging datasets: AIME 2024~\citep{veeraboina2023aime}, AIME 2025~\citep{aime25}, AMC 2023, MATH-500~\citep{hendrycks2021measuringmath500}, Minerva Math~\citep{lewkowycz2022solvingminerva}, and OlympiadBench~\citep{he2024olympiadbench}.

\paragraph{Baselines}
To validate the effectiveness of \Model{}, we compare against three distinct categories of methods:
 \textbf{Discrete CoT}: The backbone models using standard discrete Chain-of-Thought decoding without additional training.
   \textbf{Stochastic Soft Thinking}: A recent strong training-free continuous reasoning baseline~\citep{wu2025demystifyingsoftthinkinggumbelnoise}. Building on the original deterministic Soft Thinking, this method injects stochasticity via the Gumbel--Softmax trick to mitigate the “greedy pitfall” identified in the original Soft Thinking and to enable exploration at test time.
    \textbf{Discrete RL}: The backbone models fine-tuned with GRPO on the same dataset using standard discrete tokens. This serves as the direct baseline to measure the gain from continuous exploration.

\subsection{Pass@1 performance}

Table~\ref{tab:main_topp0.95} reports the Pass@1 accuracy across six mathematical reasoning benchmarks. Our proposed \Model{} consistently outperforms baselines, achieving the best results in 11 out of 12 experimental settings. 
Notably, it surpasses Discrete RL sharing the identical GRPO training setup across all the tasks. This observation validates the efficacy of exploring \model{} trajectories, proving that the performance gains stem from our unique representation brought from multiplex tokens rather than the RL training process alone. 

 Compared with Stochastic Soft Thinking, \Model{} demonstrates superior scaling behavior. On the 1.5B backbone, our method already demonstrates superior performance, achieving the better results on 4 out of 6 benchmarks. This advantage becomes more amplified at the 7B scale, where \Model{} dominantly get the highest scores across all six benchmarks. We postulate that the larger model capacity is essential for resolving the interference between superposed reasoning paths, thereby allowing the 7B model to fully leverage the exploration potential of multiplex trajectories. 

Overall, \Model{} achieves the best Pass@1 performance in 11 out of the 12 evaluation settings spanning two model sizes and six datasets. This empirical results establish \Model{} as an effective method for advancing the reasoning capabilities of large language models. 

\begin{table}[htbp]
\centering
\caption{Pass@1 accuracy on six math reasoning benchmarks averaged over 64 runs. We compare discrete CoT decoding, discrete RL fine-tuning, Stochastic Soft Thinking, and our \Model{} on DeepSeek-R1-Distill-Qwen-1.5B and DeepSeek-R1-Distill-Qwen-7B. \Model{} achieves the best performance in most setups. The best results are \textbf{bolded} and the second best results are \underline{underlined} in each column.}
\label{tab:main_topp0.95}

\resizebox{\textwidth}{!}{
\begin{tabular}{l|cccccc}
\toprule
Exp Name & AIME 2024 & AIME 2025 & AMC 2023 & MATH-500 & Minerva & OlympiadBench \\
\midrule
& \multicolumn{6}{c}{\textit{DeepSeek-R1-Distill-Qwen-1.5B}} \\ 
\midrule
Discrete CoT & 10.2 & 11.3 & 36.9 & 66.3 & 23.4 & 30.5 \\
Stochastic Soft Thinking & \underline{11.2} & \textbf{13.2} & \textbf{38.7} & \underline{66.8} & \underline{25.4} & 30.6 \\
Discrete RL & 10.5 & 12.0 & 38.4 & 66.7 & 24.3 & \underline{31.2} \\
Multiplex Thinking & \textbf{11.8} & \underline{12.8} & \textbf{38.7} & \textbf{67.5} & \textbf{26.2} & \textbf{31.3} \\
\midrule
& \multicolumn{6}{c}{\textit{DeepSeek-R1-Distill-Qwen-7B}} \\ 
\midrule
Discrete CoT & 15.7 & 16.0 & 42.4 & 71.6 & 33.3 & 35.6 \\
Stochastic Soft Thinking & \underline{20.3} & \underline{19.1} & \underline{47.9} & \underline{76.5} & \underline{37.2} & \underline{40.6} \\
Discrete  RL & 17.2 & 17.1 & 44.7 & 74.1 & 35.3 & 38.0 \\
Multiplex Thinking & \textbf{20.6} & \textbf{19.7} & \textbf{50.7} & \textbf{78.0} & \textbf{38.6} & \textbf{41.7} \\
\bottomrule
\end{tabular}
} 
\end{table}

\begin{figure*}[tb]
    \centering
    \begin{subfigure}[b]{0.24\textwidth}
        \centering
        \includegraphics[width=\textwidth,trim=21 20 22.5 10,clip]{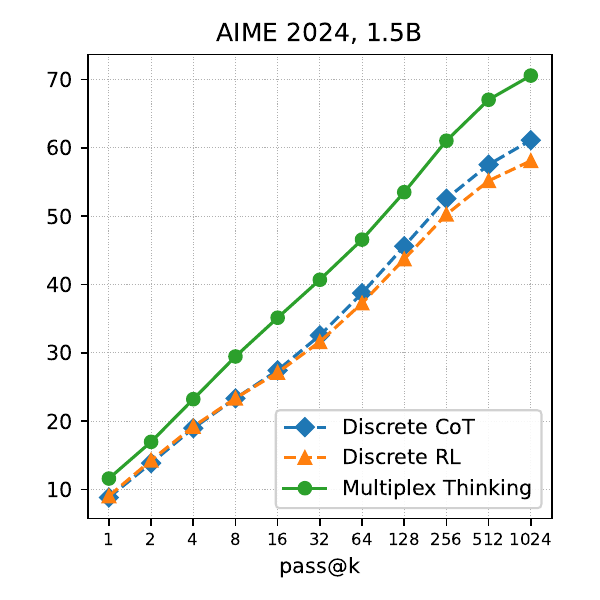}
    \end{subfigure}%
    \hfill
    \begin{subfigure}[b]{0.24\textwidth}
        \centering
        \includegraphics[width=\textwidth,trim=21 20 22.5 10,clip]{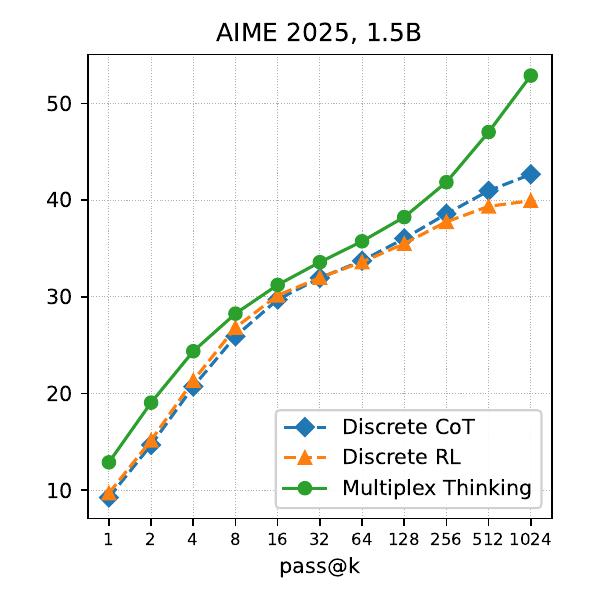}
    \end{subfigure}%
    \hfill
    \begin{subfigure}[b]{0.24\textwidth}
        \centering
        \includegraphics[width=\textwidth,trim=21 20 22.5 10,clip]{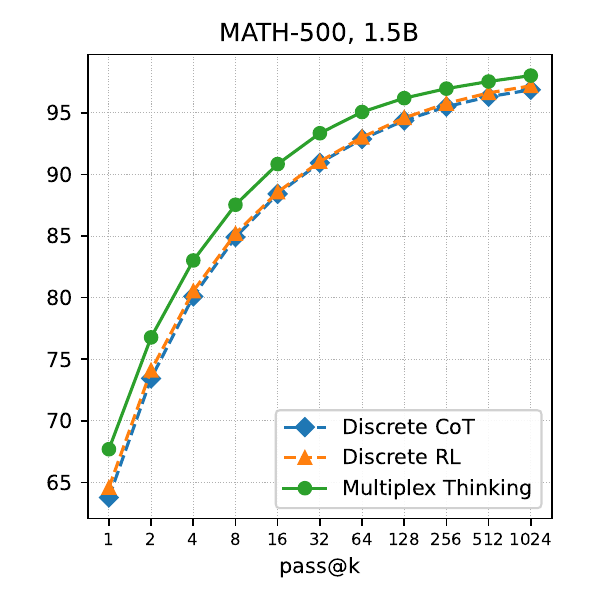}
    \end{subfigure}%
    \hfill
    \begin{subfigure}[b]{0.24\textwidth}
        \centering
        \includegraphics[width=\textwidth,trim=21 20 22.5 10,clip]{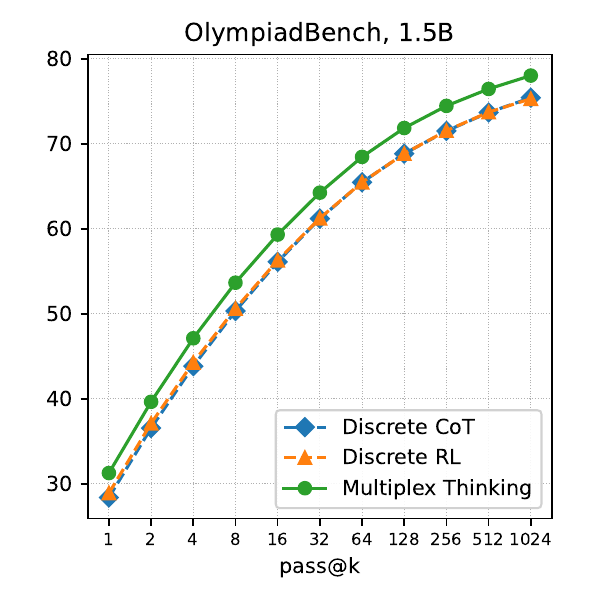}
    \end{subfigure}%
    \\ \vspace{1em} 

    \begin{subfigure}[b]{0.24\textwidth}
        \centering
        \includegraphics[width=\textwidth,trim=21 20 22.5 10,clip]{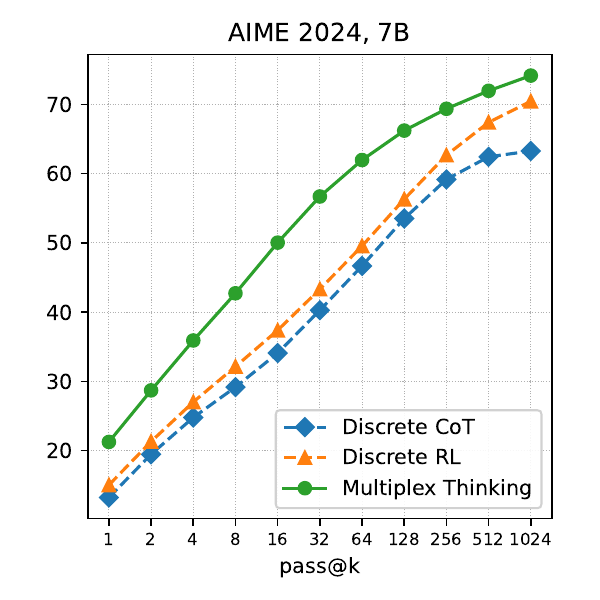}
    \end{subfigure}%
    \hfill
    \begin{subfigure}[b]{0.24\textwidth}
        \centering
        \includegraphics[width=\textwidth,trim=21 20 22.5 10,clip]{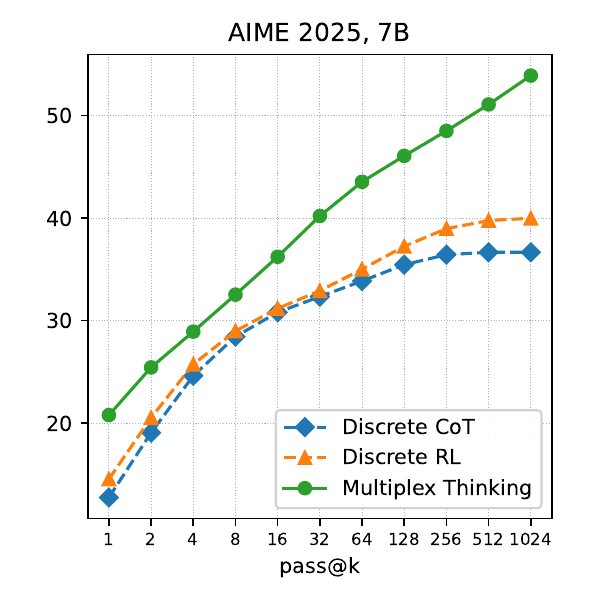}
    \end{subfigure}%
    \hfill
    \begin{subfigure}[b]{0.24\textwidth}
        \centering
        \includegraphics[width=\textwidth,trim=21 20 22.5 10,clip]{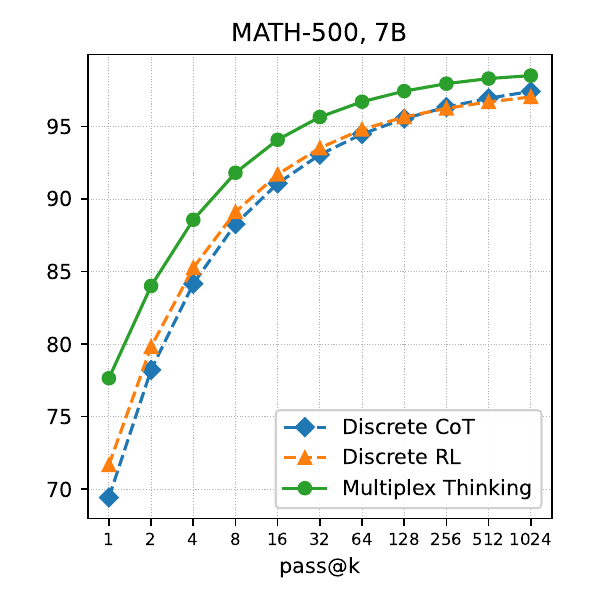}
    \end{subfigure}%
    \hfill
    \begin{subfigure}[b]{0.24\textwidth}
        \centering
        \includegraphics[width=\textwidth,trim=21 20 22.5 10,clip]{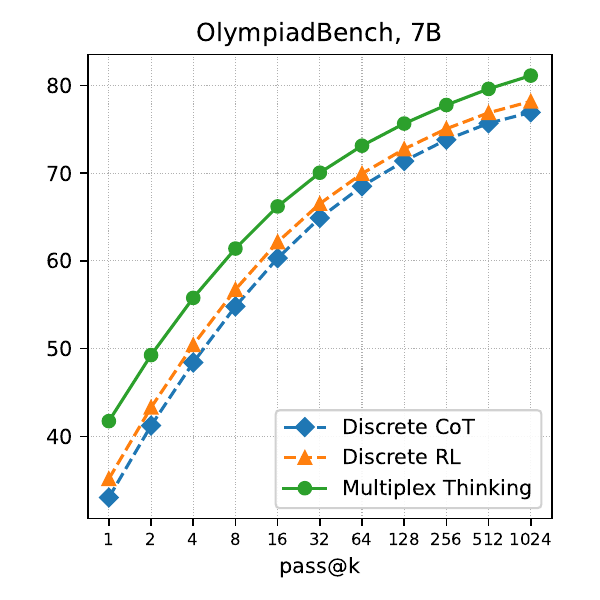}
    \end{subfigure}%
    \caption{Pass@1--Pass@1024 performance on four representative datasets. Top row: 1.5B models; bottom row: 7B models. Full results on all six datasets are reported in Appendix~\ref{app:additional_results}.}
    \label{fig:pass1024_main}
\end{figure*}

\subsection{Test-time scaling: from Pass@1 to Pass@1024}

 Pass@$k$ with a large $k$ (e.g., $k=1024$) serves as a proxy for the method's performance upper limit~\citep{yue2025doesreinforcementlearningreally}, reflecting the intrinsic exploration potential of the model. Figure~\ref{fig:pass1024_main} illustrates the Pass@$k$ performance scaling with respect to the number of sampled trajectories $k \in \{1, 2, \dots, 1024\}$.
As shown in Figure~\ref{fig:pass1024_main}, \Model{} consistently achieves a higher upper bound compared to discrete baselines in most setups.

\textbf{Exploration potential on hard tasks.} The performance gap between \Model{} and discrete baselines has a trend to wide on challenging setups as $k$ increases. For instance, on AIME 2025 (7B), while the Discrete RL baseline begins to plateau around 40\%, \Model{} continues to scale effectively, reaching approximately 55\% at $k=1024$. This substantial margin suggests that the continuous multiplex representation effectively expands the viable search space, enabling the model to uncover correct reasoning paths that are assigned negligible probability in the discrete token space. 

\textbf{Scaling behavior differs across difficulties.} We could observe from Figure~\ref{fig:pass1024_main} that the benefits of \Model{} are difficulty-dependent. On simpler datasets like MATH-500, performance saturates quickly for all methods as accuracy approaches the ceiling. However, on tasks with sparse solution spaces (e.g., AIME 2025 and OlympiadBench), the ability of multiplex tokens to maintain superposed reasoning states proves crucial for escaping local optima, resulting in the "widening gap" trend observed in the figures.

\textbf{Sampling efficiency.} Beyond higher upper limits as exploration potential, \Model{} demonstrates superior sample efficiency. To achieve a target accuracy, our method requires significantly fewer samples than discrete baselines, directly translating to reduced test-time compute.

\section{Analysis}

\begin{figure}[t]
    \centering
   \includegraphics[clip, trim={1cm 1.5cm 0cm 0cm}, width=\linewidth]{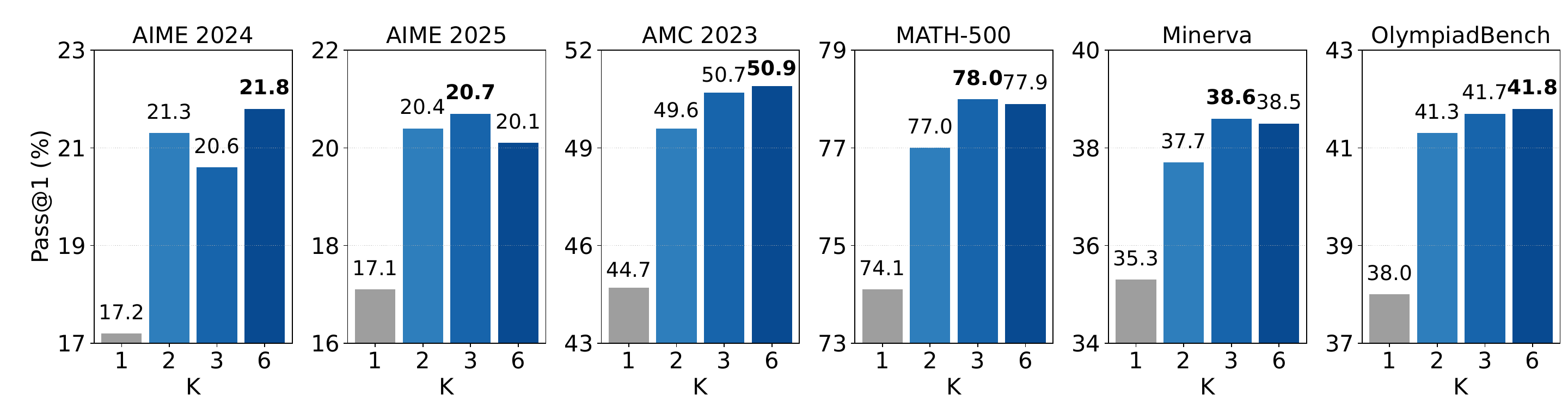}
     \caption{Performance comparison under different multiplex widths $K$. The x-axis is the multiplex width $K$ and the y-axis is the Pass@1 performance on each dataset. The grey columns ($K=1$) indicate the discrete RL performance and the blue columns represent \Model{}. The best results are \textbf{bolded} in each dataset.}
  \label{fig:k_ablation_6datasets}
\end{figure}
We analyze \Model{} through four questions that mirror the order of the following subsections. 
(1) \textbf{Do multiplex representations help without training?} Next, we evaluate an inference-only variant (\texttt{\Model{}-I}) to isolate the intrinsic benefit of multiplex trajectories from RL optimization (Table ~\ref{tab:infer_topp095_1_5b}).
(2) \textbf{How does multiplex width affect performance?} We first study the impact of multiplex width $K$ on Pass@1, with $K=1$ recovering the Discrete RL baseline (Section ~\ref{sec:ablate_on_K}). 
(3) \textbf{What is the compute trade-off between multiplex width and sequence length?} We then quantify how multiplex width can substitute for longer discrete rollouts (Table ~\ref{tab:scale_length} and Figure~\ref{fig:length_scaling}).  
(4) \textbf{Which mechanisms and design choices drive the gains?} Finally, we examine training dynamics (policy entropy and response length), ablate the token aggregation strategy, and provide qualitative visualizations to illustrate how multiplex tokens encode and modulate multiple reasoning paths in practice (Section~\ref{sec:experiment:entropy}, Section~\ref{sec:ablate_on_weight_or_unweight}, and Figure~\ref{fig:multiplex_traj}).

\subsection{Intrinsic capabilities of multiplex representation}

To disentangle the gains attributed to the multiplex representation from those yielded by Reinforcement Learning, we evaluate a training-free variant of our method, denoted as \texttt{\Model{}-I}. As shown in Table~\ref{tab:infer_topp095_1_5b},  we compare this inference-only baseline against the other two training-free baselines, including Discrete CoT and Stochastic Soft Thinking on the 7B model. Remarkably, applying \Model{} solely at inference time yields substantial performance gains over standard Discrete CoT. Compared to Stochastic Soft Thinking, \Model{}-I remains highly competitive, achieving better results on four out of six datasets. These empirical evidences demonstrate that the intrinsic capabilities of multiplex representation are beneficial for LLM reasoning. This inference-time superiority provides a strong starting point for further RL training; indeed, as shown in the final row of Table~\ref{tab:infer_topp095_1_5b}, applying RL optimization further amplifies these gains, consistently achieving the best performance across all benchmarks.

\begin{table}[tb]
\centering
\caption{Pass@1 (\%) performance conducted on 7B backbone. The best results are \textbf{bolded} and the second best results are \underline{underlined} in each column.}
\label{tab:infer_topp095_1_5b}
\resizebox{\textwidth}{!}{
\begin{tabular}{l|cccccc}
\toprule
Exp Name & AIME 2024 & AIME 2025 & AMC 2023 & MATH500 & Minerva & OlympiadBench \\
\midrule
Discrete CoT & 15.7 & 16.0 & 42.4 & 71.6 & 33.3 & 35.6 \\
Stochastic Soft Thinking & 20.3 & 19.1 & 47.9 & \underline{76.5} & \underline{37.2} & \underline{40.6} \\
\Model{}-I & \underline{20.5} & \underline{19.6} & \underline{48.6} & 76.4 & 37.1 & \underline{40.6} \\ 
\Model{} & \textbf{20.6} & \textbf{19.7} & \textbf{50.7} & \textbf{78.0} & \textbf{38.6} & \textbf{41.7} \\
\bottomrule
\end{tabular}
}
\end{table}

\subsection{The impact of token width $K$}\label{sec:ablate_on_K}
We investigate the impact of the multiplex width $K$, the number of independently sampled discrete tokens aggregated into a multiplex token. Figure~\ref{fig:k_ablation_6datasets} compares the performance trained on the 7B backbone across varying widths $K \in \{1, 2, 3, 6\}$, where $K=1$ corresponds to the standard Discrete RL baseline.

\textbf{Breaking the single-token bottleneck with $K \ge 2$.} As shown in  Figure~\ref{fig:k_ablation_6datasets}, transitioning from a single discrete token ($K=1$) to a multiplex representation ($K\ge2$) yields substantial gains across all benchmarks. And the performance gap between $K=1$ and $K \ge 2$ is significant and consistent. For example, on AMC 2023, the precision jumps from 44.7\% to 49.6\% ({+4.9\%}). This highlights that the primary advantage of our method stems from the paradigm shift which breaks the single-token bottleneck to enable exploration in a continuous latent space. 

\textbf{Diminishing marginal gains with larger $K$.} Beyond this initial leap from $K=1$ to $K=2$, we observe performance continues to increase when multiplex width $K$ increases from 2 to 3 and 6 on AMC23, MATH-500, Minerva, and OlympiadBench. However, the marginal utility of widening the multiplex window gradually diminishes. As shown in Figure~\ref{fig:k_ablation_6datasets}, the performance increase from $K=2$ to $K \in \{3, 6\}$ is notable, yet the difference between $K=3$ and $K=6$ becomes considerably smaller. This suggests that the marginal gains from exploring additional tokens are most pronounced in the initial expansion. Therefore, a moderate width (e.g., $K=3$ in our main experiments) is typically sufficient to capture the high-probability modes of the reasoning distribution, effectively covering the critical diverse paths. Further increasing the sample size yields diminishing returns, as the most valuable exploration directions are likely already included within the first few samples. We also provide Pass@$k$ for $k\in\{1,2,\dots,1024\}$ analysis in Appendix~\ref{appendix:ablate_on_K_pass_at_k}.

Beyond accuracy, we analyze how $K$ influences exploration behavior (policy entropy) and trajectory compactness (response length).

\begin{figure}[tb]
    \centering
    \small

    \begin{minipage}[t]{0.48\linewidth}
        \centering
        \captionof{table}{Pass@1 accuracy averaged on six math reasoning datasets over 64 runs. We compare \Model{}-I-4k could match the performance of Discrete CoT-5k which has 25\% more sequence token length budget.
        }
        \label{tab:scale_length}
        \begin{tabular}{l|c}
            \toprule
            Exp Name & Averaged Accuracy (\%) \\
            \midrule
            Discrete CoT-4k & 35.8 \\
            Discrete CoT-5k & 39.6 (+2.8)\\
            \Model{}-I-4k   & 40.5 (+4.7) \\
            \bottomrule
        \end{tabular}
    \end{minipage}
    \hfill
    \begin{minipage}[t]{0.48\linewidth}
        \centering
        \captionof{table}{Entropy reduction ratio (\%) under different multiplex width $K$, measured as the relative decrease in average policy entropy from the beginning to the end of training.
        }
        \label{tab:entropy_k}
        \begin{tabular}{cc}
            \toprule
            \textbf{$K$ Size} & \textbf{Entropy Reduction (\%)} \\
            \midrule
            1 (Discrete RL) & 9.44 \\
            2               & 5.82 \\
            3               & 6.03 \\
            6               & 7.09 \\
            \bottomrule
        \end{tabular}
    \end{minipage}
\end{figure}

\begin{figure}[tb]
    \centering

    \begin{minipage}[t]{0.425\linewidth}
        \centering
        \includegraphics[width=\linewidth]{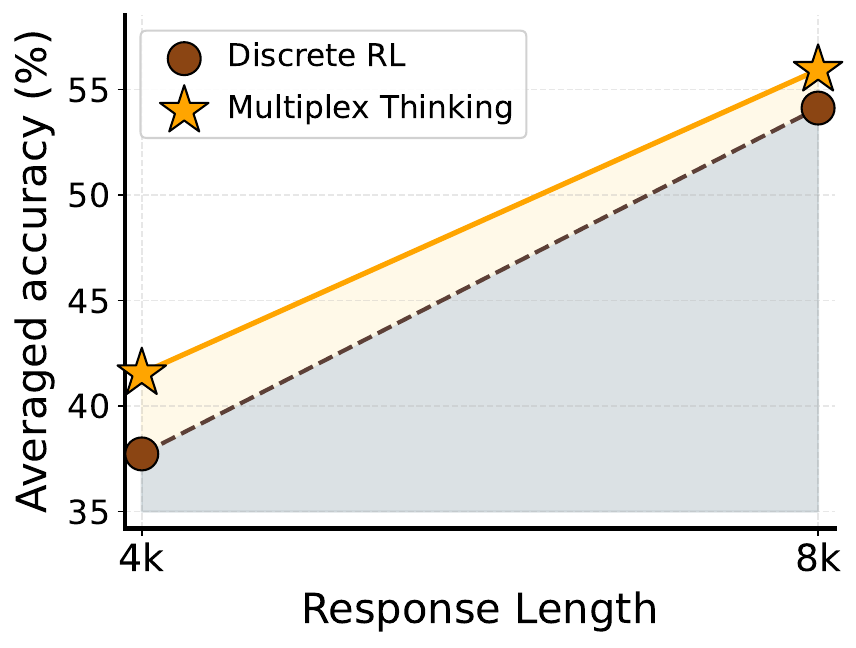}
        \vspace{-10pt}
        \captionof{figure}{Length scaling comparison. The y-axis is the accuracy averaged on six challenging reasoning datasets.}
        \label{fig:length_scaling}
    \end{minipage}
    \hfill
    \begin{minipage}[t]{0.525\linewidth}
        \centering
        \includegraphics[clip, trim={0cm 0cm 0.2cm 0cm}, width=\linewidth]{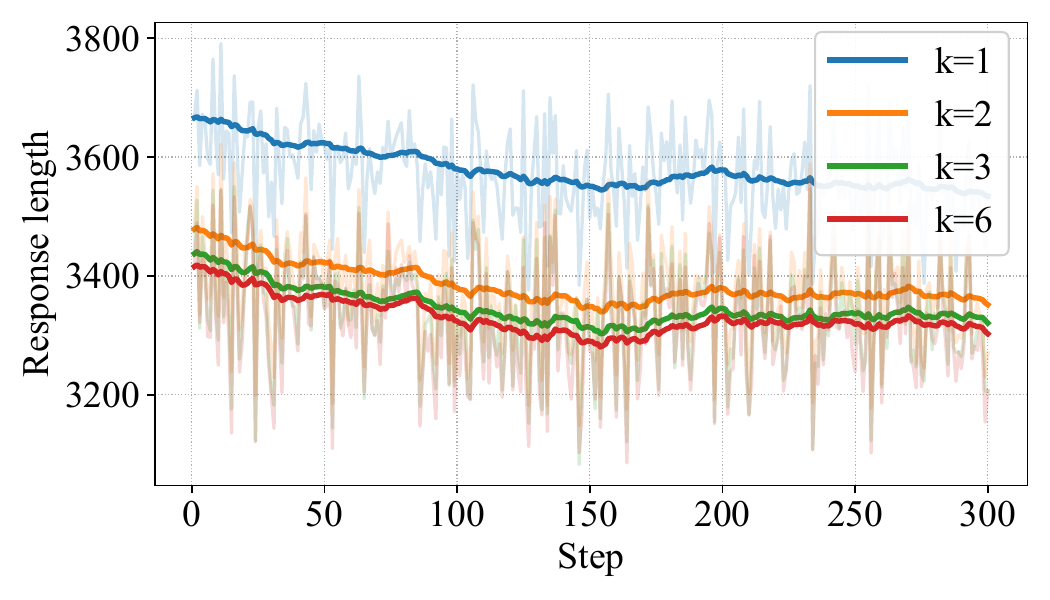}
        \vspace{-10pt}
        \captionof{figure}{Response length dynamics under different multiplex width $K$. The x-axis is the training step and the y-axis is the averaged response length at each step.}
        \label{fig:training_length}
    \end{minipage}

\end{figure}

\subsection{Test-time compute: multiplex width v.s. sequence length}
We study the trade-off between multiplex width and sequential length under different test-time compute budgets. Figure~\ref{fig:length_scaling} shows that increasing the response length improves the performance of both discrete CoT and \Model{}-I, as expected. However, even with substantially shorter trajectories, \Model{} consistently achieves higher accuracy.

To quantify this exchange rate between multiplex width and sequential length, we conduct a controlled experiment by scaling the inference token budget of the discrete CoT baseline from 4,096 to 5,120, while strictly constraining \Model{}-I to a 4,096-token budge, denoted as \Model{}-I-4k.
As shown in Table~\ref{tab:scale_length}, the performance of discrete CoT increases with raised sequence length budget. However, \Model{}-I-4k with a 4k limit consistently outperforms or matches the Discrete CoT-5k baseline with 20\% shorter sequences. 
This indicates that performance gains do not solely depend on longer discrete rollouts, but can instead be achieved through richer token representations.

Further evidence is provided by the training dynamics in Figure~\ref{fig:training_length},where we observe that \Model{} consistently generates trajectories with fewer tokens than the Discrete RL baseline, yet achieves superior accuracy. This observation aligns the intuition of \model{} that multiplex tokens possess higher information density. Since each token encodes multiple potential paths, the model can express complex reasoning steps more compactly, effectively compressing the reasoning process in a shorter response.  Importantly, increasing $K$ changes the number of sampled candidates per step, but does not require additional forward passes beyond sampling from the same logits distribution.

\subsection{Entropy analysis}
\label{sec:experiment:entropy}

We analyze how multiplex tokens influence exploration during RL by measuring the \emph{entropy reduction ratio} of the policy. 
Concretely, for each setting we compute $H_{\text{start}}$ as the average policy entropy over the first 10 training steps and $H_{\text{end}}$ over the last 10 steps, and report $(H_{\text{start}}-H_{\text{end}})/H_{\text{start}} \times 100$ (Table~\ref{tab:entropy_k}). 
A smaller reduction ratio indicates less entropy collapse and thus more sustained exploration throughout training.

As shown in Table~\ref{tab:entropy_k}, multiplex training exhibits consistently lower entropy reduction than the discrete RL baseline ($K=1$), suggesting that multiplex tokens mitigate premature commitment to a single reasoning path. 
This trend is consistent with the higher Pass@k upper bounds observed for larger $K$ (Figure~\ref{fig:pass1024_main}), where maintaining exploration helps discover correct trajectories that would otherwise receive negligible probability under discrete decoding.

\subsection{Ablation study on token aggregation strategy}\label{sec:ablate_on_weight_or_unweight}
We conduct ablation study on a crucial component of \model{}: the token aggregation strategy.
We investigate the impact of the aggregation strategy used to construct the multiplex token representation. Specifically, we compare the probability-weighted summation (denoted as \Model{}-Weighted) against a simple unweighted average (denoted as \Model{}-Averaged) of the $K$ token embeddings. As detailed in Table~\ref{tab:ablate_on_embed}, both strategies yield highly comparable empirical performance across model scales and benchmarks, and both consistently outperform the Discrete RL baseline by a significant margin. The similarity suggests that the effectiveness of \Model{} stems from the inclusion of diverse reasoning paths in the latent space rather than the specific weighting scheme used to combine them. This finding highlights the robustness of our approach, as the model can effectively learn to extract relevant features from the multiplex representation regardless of the precise linear combination coefficients.

\begin{table}[htbp]
\centering
\caption{Pass@1 accuracy on six math reasoning datasets, averaged over 64 runs. We compare discrete RL and our \Model{} with two different token strategies, denoted as \Model{}-Averaged and \Model{}-Weighted respectively. The best results are \textbf{bolded}.}
\label{tab:ablate_on_embed}
\resizebox{\textwidth}{!}{
\begin{tabular}{l|cccccc}
\toprule
Exp Name & AIME 2024 & AIME 2025 & AMC 2023 & MATH-500 & Minerva & OlympiadBench \\
\midrule
& \multicolumn{6}{c}{\textit{DeepSeek-R1-Distill-Qwen-1.5B}} \\ 
\midrule
Discrete RL & 10.5 & 12.0 & 38.4 & 66.7 & 24.3 & 31.2 \\
Multiplex Thinking-Averaged & 11.7 & \textbf{14.2} & 38.1 & \textbf{67.7} & \textbf{26.3} & 31.0 \\
Multiplex Thinking-Weighted & \textbf{11.8} & 12.8 & \textbf{38.7} & 67.5 & 26.2 & \textbf{31.3} \\
\midrule
& \multicolumn{6}{c}{\textit{DeepSeek-R1-Distill-Qwen-7B}} \\ 
\midrule
Discrete  RL & 17.2 & 17.1 & 44.7 & 74.1 & 35.3 & 38.0 \\
Multiplex Thinking-Averaged & 19.9 & \textbf{20.0} & 49.9 & 77.6 & 38.4 & \textbf{41.8} \\
Multiplex Thinking-Weighted & \textbf{20.6} & 19.7 & \textbf{50.7} & \textbf{78.0} & \textbf{38.6} & 41.7 \\
\bottomrule
\end{tabular}
}
\end{table}

\subsection{Qualitative Analysis}

\begin{figure}[tb]
    \centering
    \begin{tokenforkbox}
  \tok{darkblue}{ I}\tok{lightblue}{ The} need to figure out how many different ways\tok{darkblue}{ the}\tok{lightblue}{ these} six cars can stack up\tok{darkblue}{ at}\tok{lightblue}{ so} the intersection\tok{darkblue}{ such}\tok{lightblue}{ so} that all three lanes are\tok{darkblue}{ occupied}\tok{lightblue}{ used}.\tok{darkblue}{ The}\tok{lightblue}{\textbackslash{}n}\tok{lightestblue}{ Hmm} cars are distinguishable,\tok{darkblue}{ meaning}\tok{lightblue}{ and} each\tok{darkblue}{ one}\tok{lightblue}{ car} is unique, and\tok{darkblue}{ the}\tok{lightblue}{ they}\tok{lightestblue}{ since} order in which they\tok{darkblue}{ approach}\tok{lightblue}{ arrive}\tok{darkblue}{ matters}\tok{lightblue}{ is} fixed\tok{darkblue}{.}\tok{lightblue}{.\textbackslash{}n\textbackslash{}n}\tok{darkblue}{ So}\tok{lightblue}{\textbackslash{}n}\tok{lightestblue}{ Hmm},\tok{darkblue}{ the}\tok{lightblue}{ car}\tok{lightestblue}{ different}\tok{darkblue}{ first}\tok{lightblue}{ cars} car\tok{darkblue}{ arrives}\tok{lightblue}{ gets}\tok{darkblue}{ first}\tok{lightblue}{,}, then the second, and so on\tok{darkblue}{ up}\tok{lightblue}{ until} to the sixth\tok{darkblue}{.}\tok{lightblue}{ car}\textbackslash{}n\tok{darkblue}{Hmm}\tok{lightblue}{First}.\tok{darkblue}{ Let}\tok{lightblue}{ So}\tok{darkblue}{ me}\tok{lightblue}{'s}\tok{darkblue}{ try}\tok{lightblue}{ break}\tok{lightestblue}{ start} to\tok{darkblue}{ break}\tok{lightblue}{ parse} this down.\tok{darkblue}{ Since}\tok{lightblue}{ First}\tok{darkblue}{ the}\tok{lightblue}{ each} car has three choices,%
\tok{darkblue}{ without}\tok{lightblue}{ if}\tok{lightestblue}{ I}\tok{darkblue}{ might}\tok{lightblue}{ could}\tok{lightestblue}{ guess} think\tok{darkblue}{ at}\tok{lightblue}{ initially} first\tok{darkblue}{ that}\tok{lightblue}{ it}\tok{darkblue}{ there}\tok{lightblue}{ the} total number of\tok{darkblue}{ possible}\tok{lightblue}{ ways}\tok{darkblue}{ ways}\tok{lightblue}{ arrangements} they can\tok{darkblue}{ choose}\tok{lightblue}{ stack}\tok{lightestblue}{ move}\tok{darkblue}{ their}\tok{lightblue}{ the} lanes is\tok{darkblue}{ }\tok{lightblue}{ just}3\textasciicircum{}6\tok{darkblue}{.}\tok{lightblue}{,}\tok{darkblue}{ But}\tok{lightblue}{ Let}\tok{lightestblue}{ Because} wait,\tok{darkblue}{ that}\tok{lightblue}{ the}\tok{darkblue}{ would}\tok{lightblue}{'s}\tok{lightestblue}{ includes}\tok{darkblue}{ be}\tok{lightblue}{ include} the case\tok{darkblue}{ without}\tok{lightblue}{ if}\tok{darkblue}{ the}\tok{lightblue}{ all}\tok{lightestblue}{ they}\tok{darkblue}{ order}\tok{lightblue}{ problem}\tok{darkblue}{ didn}\tok{lightblue}{ of}\tok{lightestblue}{ in}'t matter\tok{darkblue}{,}\tok{lightblue}{ or}\tok{darkblue}{ but}\tok{lightblue}{ right}\tok{darkblue}{ here}\tok{lightblue}{ since}\tok{darkblue}{,}\tok{lightblue}{ the}\tok{darkblue}{ since}\tok{lightblue}{ the}\tok{darkblue}{ order}\tok{lightblue}{ cars}\tok{darkblue}{ does}\textcolor{lightblue}{ in} matter because\tok{darkblue}{ the}\tok{lightblue}{ each} cars\tok{darkblue}{ are}\tok{lightblue}{ pull}\tok{darkblue}{ distinguish}\tok{lightblue}{ coming}able and\tok{darkblue}{ they}\tok{lightblue}{ arrive}\tok{lightestblue}{ pull}\tok{darkblue}{ arrive}\tok{lightblue}{ pull}\tok{lightestblue}{ come}\tok{darkblue}{ one}\tok{lightblue}{ sequentially}\tok{darkblue}{ after}\tok{lightblue}{ at} another.\tok{darkblue}{ So}\tok{lightblue}{ Hmm}\tok{lightestblue}{ However}\tok{darkblue}{,}\tok{lightblue}{ maybe}\tok{darkblue}{ maybe}\tok{lightblue}{ that}\tok{darkblue}{ each}\tok{lightblue}{ considering}\tok{darkblue}{ arrangement}\tok{lightblue}{ car} is\tok{darkblue}{ a}\tok{lightblue}{ just}\tok{darkblue}{ permutation}\tok{lightblue}{ sequence}\tok{darkblue}{ with}\tok{lightblue}{ where}\tok{darkblue}{ repetition}\tok{lightblue}{ specific}\tok{darkblue}{?}\tok{lightblue}{?\textbackslash{}n\textbackslash{}n}\tok{darkblue}{ So}\tok{lightblue}{\textbackslash{}n}\tok{darkblue}{,}\tok{lightblue}{ yes}\tok{darkblue}{ yeah}\tok{lightblue}{ thinking},\tok{darkblue}{ so}\tok{lightblue}{ }\tok{lightestblue}{ I}\tok{darkblue}{ each}\tok{lightblue}{ }\tok{lightestblue}{ that}\tok{darkblue}{ total}\tok{lightblue}{ each}\tok{lightestblue}{ three}\tok{darkblue}{ number}\tok{lightblue}{ without}\tok{darkblue}{ of}\tok{lightblue}{ is}\tok{lightestblue}{ should}\tok{darkblue}{ ways}\tok{lightblue}{ possible}\tok{lightestblue}{ possibilities}\tok{darkblue}{ without}\tok{lightblue}{ is}\tok{lightestblue}{ should}\tok{darkblue}{ any}\tok{lightblue}{ the} restrictions\tok{darkblue}{ would}\tok{lightblue}{ is}\tok{darkblue}{ be}\tok{lightblue}{ indeed} 3\tok{darkblue}{ multiplied}\tok{lightblue}{\textasciicircum{}}6\tok{darkblue}{,}\tok{lightblue}{.} which is 729\tok{darkblue}{.\textbackslash{}n\textbackslash{}n}\tok{lightblue}{.} 
\end{tokenforkbox}
\caption{\textbf{Visualization of a representative \model{}trajectory ($K=3$).} At each step, the model samples $K=3$ independent discrete tokens. The boxes summarize the sample outcomes:(i) when all three samples are distinct, the tokens are shown with \tokDarkBlueBox{yellow}, \tokLightBlueBox{purple}, and \tokLightestBlueBox{red} boxes to indicate high diversity; (ii) when two samples agree, a \tokDarkBlueBox{yellow box} marks the majority token (sampled twice) and a \tokLightBlueBox{purple box} marks the minority token (sampled once); (iii) plain text (no box) indicates complete consensus, where all three samples are identical.
}
\label{fig:multiplex_traj}
\end{figure}

To better understand how \Model{} operates in practice, we visualize a representative reasoning trajectory in Figure~\ref{fig:multiplex_traj}. The figure displays a representative multiplex trajectory while solving a math problem. A key observation is that \Model{} effectively modulates its exploration strategy based on the uncertainty of the current reasoning state. Concretely, the trajectory alternates between \emph{consensus} and \emph{exploration} phases. 
During \emph{consensus} steps (unboxed), the sampled candidates collapse to the same token, indicating a peaked next-token distribution and a locally stable reasoning state. In contrast, during \emph{exploration steps} highlighted by \tokDarkBlueBox{yellow}, \tokLightBlueBox{purple}, and \tokLightestBlueBox{red} where divergent candidates are sampled, \Model{} compacts these alternatives into a single continuous multiplex token and continues the rollout while preserving uncertainty. 

Notably, the highlighted exploration steps correspond to higher-entropy positions where multiple plausible continuations compete. \Model{} explicitly retains these alternatives via multiplex aggregation, enabling branching behavior at the very tokens that act as decision points. This is consistent with recent findings that high-entropy minority tokens function as critical forks in CoT reasoning and account for most gains in RLVR~\citep{wang20258020rulehighentropyminority}.
We also provide a full trajectory in Appendix~\ref{app:full_traj}.

\section{Related Works}

\noindent \textbf{Discrete Reasoning} Chain-of-Thought (CoT) prompting~\citep{wei2022chain} has become the standard training-free method to elicit reasoning in LLMs. 
To further enhance these capabilities, methods like STaR~\citep{zelikman2022star} and ReST~\citep{gulcehre2023reinforced} utilize iterative fine-tuning on self-generated rationales. 
More recently, DeepSeek-R1~\citep{deepseekr1} demonstrated that large-scale RL with verifiable rewards can incentivize the reasoning capability in LLMs. However, discrete RL methods suffer from high computation cost of generating long sequences by conducting a depth-first style decoding to reach solutions~\citep{zhu2025reasoningsuperpositiontheoreticalperspective}.

\noindent \textbf{Continuous Reasoning} 
A line of studies~\citep{yang2024latentmultihop,biran2024hoppinglateexploringlimitations,hao2025coconut} define the hidden states of the transformer as latent reasoning steps.
\citet{yang2024latentmultihop} showed intermediate reasoning steps can be decoded from transformer hidden states. COCONUT~\citep{hao2025coconut} trained Large Language Models (LLMs) by using the last hidden states of the transformer as input embeddings and showed continuous chain of thought (CoT) could outperform discrete CoT on logical reasoning tasks.
However, using hidden states as embeddings can suffer from representational misalignment~\citep{softthinking} when input embeddings and the prediction head become decoupled in larger LLMs, and full-model retraining can induce catastrophic forgetting~\citep{xu2025softcot}.
Another line of study\citep{softthinking} proposed to utilize probability-weighted mixture of token embeddings as continuous tokens. This preserves the embedding prior and avoid extensive retraining.Both families of continuous tokens are inherently deterministic. 
Some recent works attempt to introduce stochasticity into continuous reasoning by injecting external noise Gaussian noise~\citep{butt2025softtokenshardtruths} or Gumbel noise~\citep{wu2025demystifyingsoftthinkinggumbelnoise} into the logits. However, none of these methods have explored using multiple independent sampling to form a single aggregated token representation at each decoding step that could achieve the best of discrete sampling and continuous representation. 

\textbf{Parallel Reasoning} 
Parallel reasoning improves problem solving by exploring multiple reasoning paths and aggregating their outcomes.
Representative methods include self-consistency~\citep{wang2023selfconsistency} and Best-of-N (BoN) selection using outcome rewards~\citep{cobbe2021training} or process rewards~\citep{lightman2023lets}, as well as search-based approaches that branch over intermediate steps such as such as Tree-of-Thought~\citep{yao2023treethoughtsdeliberateproblem} and adaptive parallel reasoning methods~\citep{pan2025learningadaptiveparallelreasoning, lian2025threadweaveradaptivethreadingefficient}.
A common drawback of parallel reasoning methods is that computation scales roughly linearly with the number of sampled paths, since each path requires generating a full sequence. 
Our proposed \model{} serves as a complementary dimension to existing parallel reasoning strategies because it changes the per-step token distribution rather than the outer-loop sampling budget. It can be seamlessly integrated into frameworks like Self-Consistency or BoN to further push the boundaries of reasoning.

\section{Conclusion}
In this work, we introduced \Model{}, a novel framework that bridges the gap between discrete Chain-of-Thought and continuous reasoning representations. By aggregating multiple independent discrete tokens in continuous representations, our method achieves higher information density while preserving the probabilistic sampling required for effective reinforcement learning. Empirical evaluations across challenging mathematical benchmarks demonstrate that \Model{} consistently outperforms strong discrete CoT and RL baselines from Pass@1 to Pass@1024. Crucially, our analysis reveals that this performance gain comes with improved token efficiency, as the model learns to compress complex reasoning steps into shorter trajectories. These findings suggest that \model{} is a promising direction for scaling test-time compute, offering a scalable path toward more capable and efficient reasoning models.

\section*{Acknowledgements}
This work is partially supported by a gift from AWS to Penn Engineering’s ASSET Center for Trustworthy AI. We thank Junli Wang for helpful discussions on adapting SGLang. We also thank Jian Chen for insightful discussions during the development of this work. Finally, we thank Long Lian, Yonghyun Park, Junyi Zhang, and Mutian Tong for proofreading and valuable feedback on the manuscript.

\bibliography{iclr2026_conference}
\bibliographystyle{iclr2026_conference}
\newpage
\appendix

\section{Appendix}

\subsection{Implementation details}\label{Appendix:experiment_details}

We implement our framework based on verl~\citep{verl} and SGLang. Our codebase references the implementation design of Soft Thinking~\citep{softthinking}. We use the the version 0.4.9.post6 of SGLang. All experiments are conducted on 8$\times$ NVIDIA DGX B200 GPUs\footnote{Computations are conducted using bfloat16 precision on the NVIDIA Blackwell architecture to balance numerical stability and computational throughput.}.

\noindent\textbf{Stopping Criteria.} In \model{}, the transition from the thinking phase to the final answer generation is triggered when the discrete token with the highest probability corresponds to the special token \texttt{</think>}. We explicitly adopt this strategy over training-free heuristics, such as monitoring consecutive low-entropy tokens for early stopping in other works. We observed that such heuristics introduce artificial patterns that the model tends to exploit during RL optimization, resulting in training instability and the generation of incoherent content. By removing these handcrafted constraints, we allow the RL objective to naturally regulate the thinking process.
Since indefinite thinking without producing a final answer yields low rewards, the model autonomously learns to generate the stop token at appropriate steps to maximize the return.
This approach helps reduce the reward hacking associated with rule-based stopping criteria.

\subsubsection{hyper-parameters}
Table~\ref{tab:hyper-params} provides a detailed summary of the hyper-parameters used for GRPO training of the discrete RL and \model{}.

\begin{table}[h]
\centering
\begin{tabular}{lc}
\toprule
\textbf{Params} & \textbf{Values} \\
\midrule
Batch size & {128} \\
PPO mini batch size & 128 \\
Rollout number & 8 \\
Learning rate & $10^{-6}$ \\
Sampling temperature & 1.0 \\
Sampling top $p$ & 1.0 \\
Multiplex width $K$ & 3 \\
Max prompt length & 1024 \\
Max response length & 4096 \\
Entropy loss coefficient & 0 \\
KL loss coefficient & 0 \\
Model data type & bfloat16 \\
\bottomrule
\\
\end{tabular}
\caption{Hyper-parameters used in GRPO training.}
\label{tab:hyper-params}
\end{table}


\subsection{Additional results}\label{app:additional_results}

In this section, we present the complete experimental results that were omitted from the main body due to space constraints.

\subsubsection{Full Pass@1--Pass@1024 results}

\begin{figure*}[tb]
    \centering
    \begin{subfigure}[b]{0.33\textwidth}
        \centering
        \includegraphics[width=\textwidth,trim=21 20 22.5 10,clip]{iclr2026/figs/0112/passk_AIME_2024_1.5B.pdf}
    \end{subfigure}%
    \begin{subfigure}[b]{0.33\textwidth}
        \centering
        \includegraphics[width=\textwidth,trim=21 20 22.5 10,clip]{iclr2026/figs/0112/passk_AIME_2025_1.5B.pdf}
    \end{subfigure}%
    \begin{subfigure}[b]{0.33\textwidth}
        \centering
        \includegraphics[width=\textwidth,trim=21 20 22.5 10,clip]{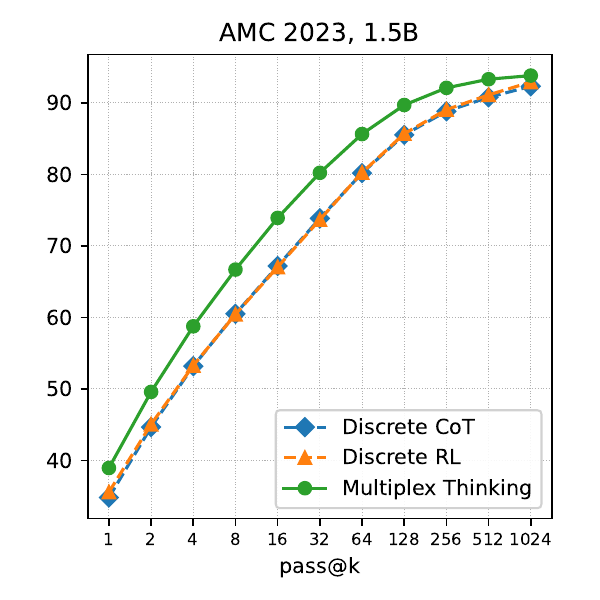}
    \end{subfigure}

    \begin{subfigure}[b]{0.33\textwidth}
        \centering
        \includegraphics[width=\textwidth,trim=21 20 22.5 10,clip]{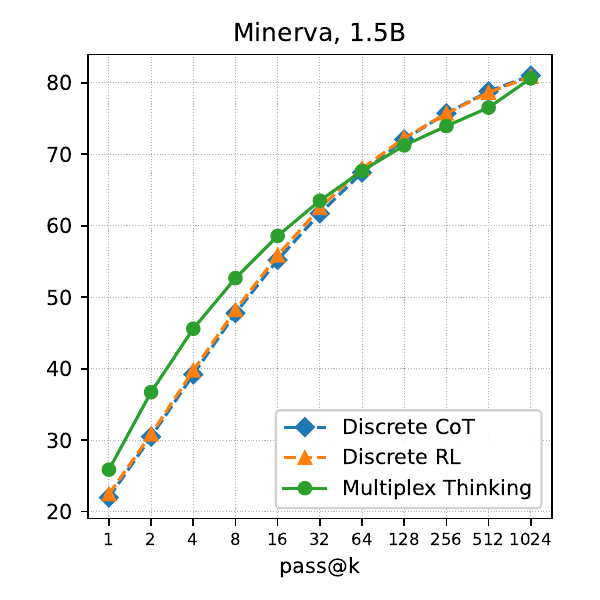}
    \end{subfigure}%
    \begin{subfigure}[b]{0.33\textwidth}
        \centering
        \includegraphics[width=\textwidth,trim=21 20 22.5 10,clip]{iclr2026/figs/0112/passk_MATH-500_1.5B.pdf}
    \end{subfigure}%
    \begin{subfigure}[b]{0.33\textwidth}
        \centering
        \includegraphics[width=\textwidth,trim=21 20 22.5 10,clip]{iclr2026/figs/0112/passk_OlympiadBench_1.5B.pdf}
    \end{subfigure}

    \begin{subfigure}[b]{0.33\textwidth}
        \centering
        \includegraphics[width=\textwidth,trim=21 20 22.5 10,clip]{iclr2026/figs/0112/passk_AIME_2024_7B.pdf}
    \end{subfigure}%
    \begin{subfigure}[b]{0.33\textwidth}
        \centering
        \includegraphics[width=\textwidth,trim=21 20 22.5 10,clip]{iclr2026/figs/0112/passk_AIME_2025_7B.pdf}
    \end{subfigure}%
    \begin{subfigure}[b]{0.33\textwidth}
        \centering
        \includegraphics[width=\textwidth,trim=21 20 22.5 10,clip]{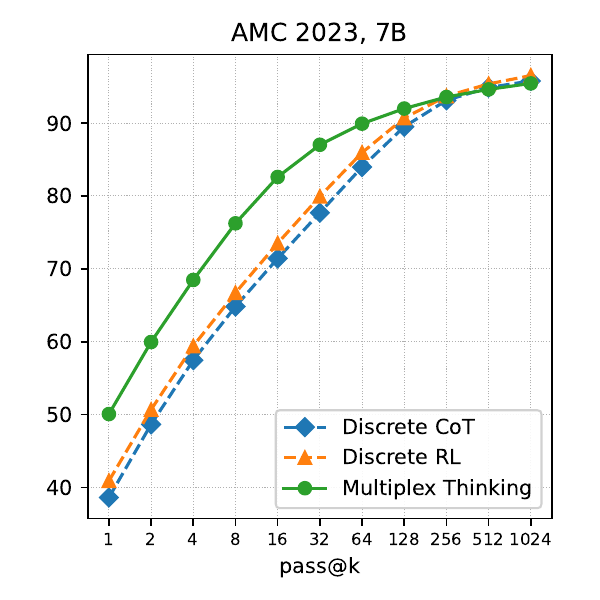}
    \end{subfigure}

    \begin{subfigure}[b]{0.33\textwidth}
        \centering
        \includegraphics[width=\textwidth,trim=21 20 22.5 10,clip]{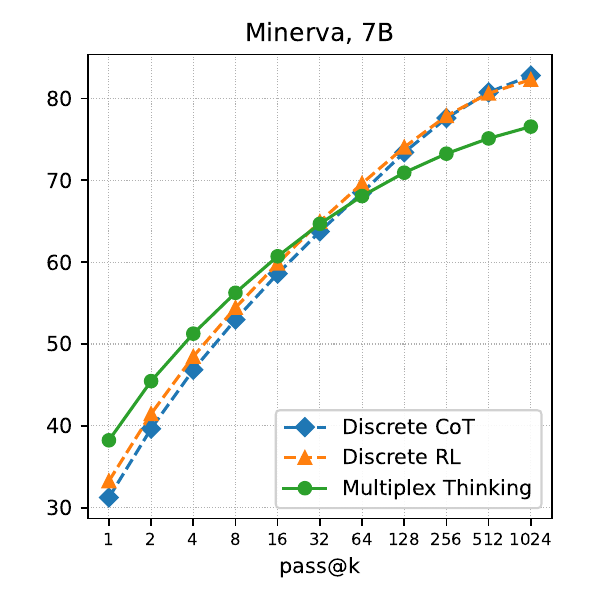}
    \end{subfigure}%
    \begin{subfigure}[b]{0.33\textwidth}
        \centering
        \includegraphics[width=\textwidth,trim=21 20 22.5 10,clip]{iclr2026/figs/0112/passk_MATH-500_7B.pdf}
    \end{subfigure}%
    \begin{subfigure}[b]{0.33\textwidth}
        \centering
        \includegraphics[width=\textwidth,trim=21 20 22.5 10,clip]{iclr2026/figs/0112/passk_OlympiadBench_7B.pdf}
    \end{subfigure}
    \caption{Pass@1 to Pass@1024 performance spanning all tasks and all models.}
    \label{fig:pass1024_appendix_whole}
\end{figure*}

Figure~\ref{fig:pass1024_appendix_whole} illustrates the detailed Pass@$k$ performance (from $k=1$ to $1024$) across all six benchmarks for both the 1.5B and 7B model scales, supplementing the representative results discussed in the main text.

\subsubsection{Pass@1--Pass@1024 results comparing different multiplex width $K$}\label{appendix:ablate_on_K_pass_at_k}

\begin{figure*}[t]
    \centering
    \begin{subfigure}[b]{0.38\textwidth}
        \centering
        \includegraphics[width=\textwidth,trim=2 20 2 10,clip]{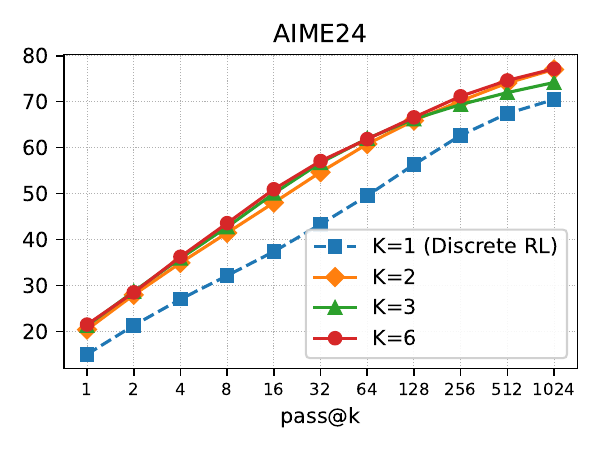}
    \end{subfigure}%
    \begin{subfigure}[b]{0.38\textwidth}
        \centering
        \includegraphics[width=\textwidth,trim=2 20 2 10,clip]{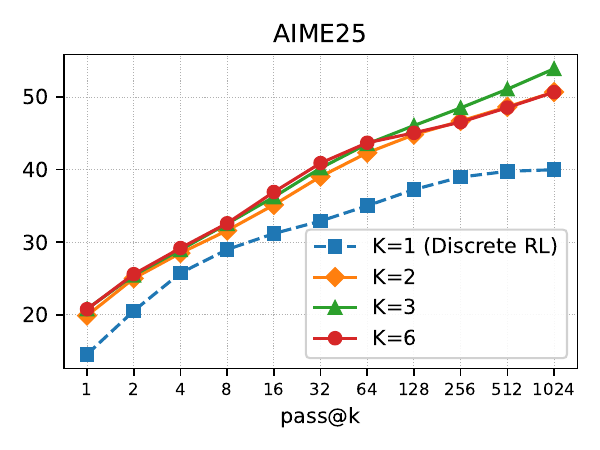}
    \end{subfigure}

    \begin{subfigure}[b]{0.38\textwidth}
        \centering
        \includegraphics[width=\textwidth,trim=2 20 2 10,clip]{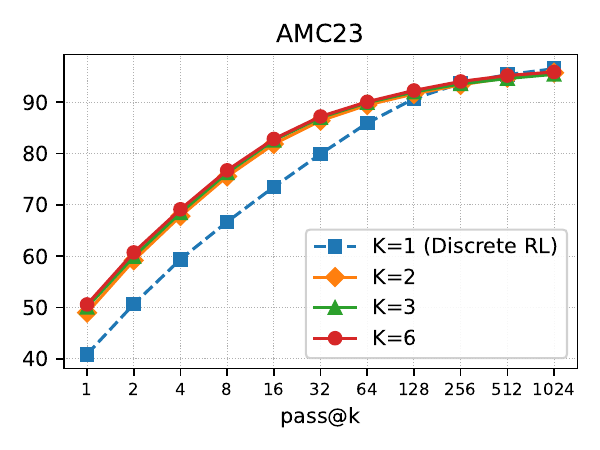}
    \end{subfigure}%
    \begin{subfigure}[b]{0.38\textwidth}
        \centering
        \includegraphics[width=\textwidth,trim=2 20 2 10,clip]{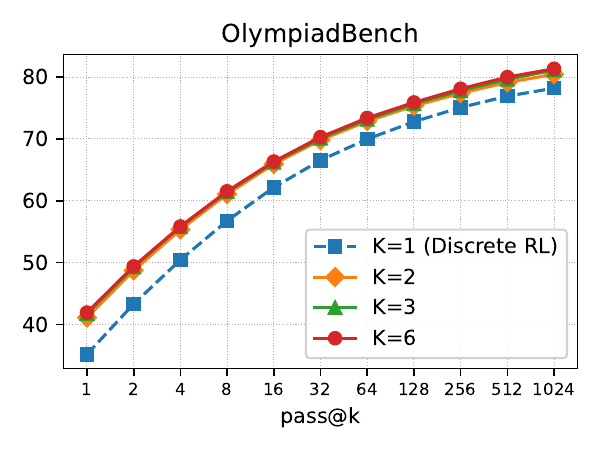}
    \end{subfigure}

    \caption{K size ablation.}
    \label{fig:ablate_on_k}
\end{figure*}

In Section~\ref{sec:ablate_on_K}, we report the ablation study on multiplex width $k$ with the Pass@1 performance using top-p of 0.95. We extend this analysis by presenting the full Pass@$k$ trajectories for $k\in\{1,2,4,\dots,1024\}$ in Figure~\ref{fig:ablate_on_k} on four representative datasets. 

From Figure~\ref{fig:ablate_on_k}, we could observe that there is a significant performance gap between the baseline Discrete RL ($K=1$) and the \model{} variants ($K \ge 2$). This empirical observation strongly corroborates our analysis in Section~\ref{sec:ablate_on_K}. While the transition from discrete RL to \model{} overcomes the limitation of depth-first style search, scaling the multiplex width further results in diminishing returns, with the performance curves for $K=2, 3,$ and $6$ remaining closely clustered.

\subsection{Training dynamics}
\begin{figure}[tb]
    \centering
    \begin{subfigure}[t]{0.48\linewidth} 
        \centering
        \includegraphics[clip, trim={0cm 0cm 0cm 0cm}, width=\linewidth]{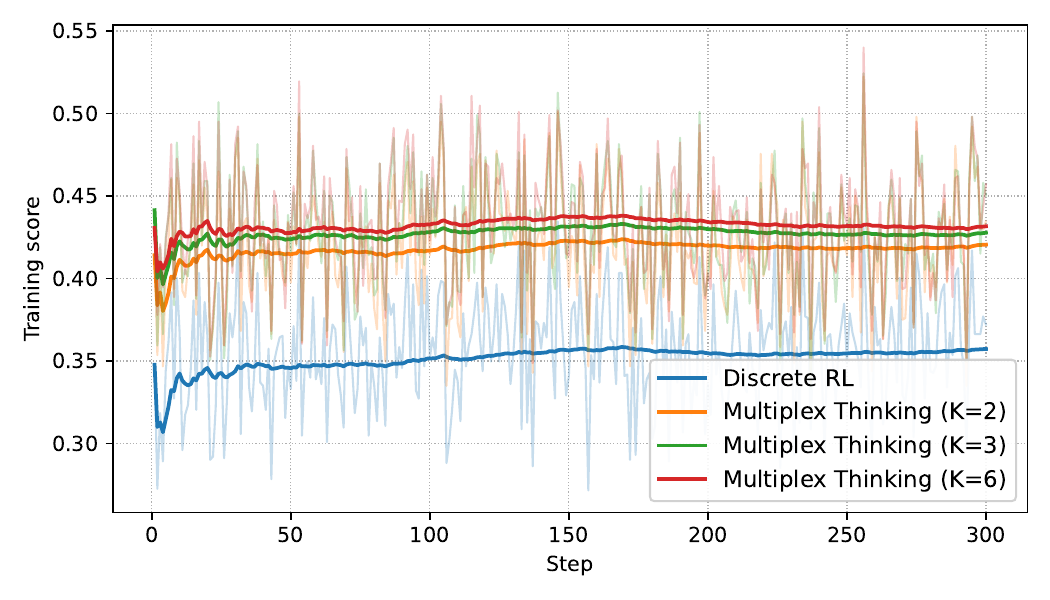}
        \caption{Training score dynamics}
        \label{fig:sub1}
    \end{subfigure}
    \begin{subfigure}[t]{0.48\linewidth}
        \centering
        \includegraphics[clip, trim={0cm 0cm 0cm 0cm}, width=\linewidth]{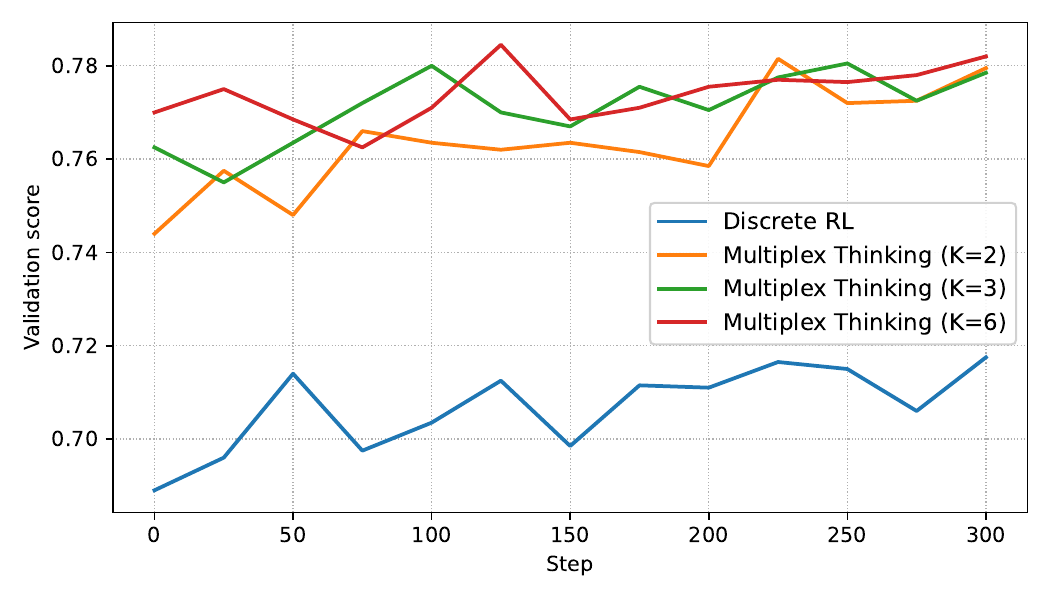}
        \caption{Validation score}
        \label{fig:sub2}
    \end{subfigure}
     \caption{Training dynamics comparison of \model{} and discrete reasoning. }
    \label{fig:training_reward}
\end{figure}

We provide the training score and validation score in Figure~\ref{fig:training_reward}. The trained model is validated every 25 training steps and we use the Pass@4 metric on MATH-500 for validation.

\subsection{Full trajectory example}\label{app:full_traj}
To qualitatively illustrate the on-policy exploration of \Model{}, we visualize a full reasoning trajectory below. 
    
\begin{tokenforkbox}
Alright,\tok{darkblue}{ so}\tok{lightblue}{ let} I\tok{darkblue}{ have}\tok{lightblue}{'ve}\tok{lightestblue}{'m} this problem\tok{darkblue}{ where}\tok{lightblue}{ here} six cars\tok{darkblue}{ are}\tok{lightblue}{ come}\tok{darkblue}{ approaching}\tok{lightblue}{ pulling}\tok{lightestblue}{ arriving} a red light\tok{darkblue}{ and}\tok{lightblue}{,}\tok{darkblue}{ and}\tok{lightblue}{ each} they\tok{darkblue}{ can}\tok{lightblue}{ have}\tok{darkblue}{ choose}\tok{lightblue}{ each}\tok{darkblue}{ between}\tok{lightblue}{ to}\tok{lightestblue}{ one} three\tok{darkblue}{ lanes}\tok{lightblue}{ different}: left-turn, straight-going, or right-turn.\tok{darkblue}{ I}\tok{lightblue}{ The} need to figure out how many different ways\tok{darkblue}{ the}\tok{lightblue}{ these} six cars can stack up\tok{darkblue}{ at}\tok{lightblue}{ so} the intersection\tok{darkblue}{ such}\tok{lightblue}{ so} that all three lanes are\tok{darkblue}{ occupied}\tok{lightblue}{ used}.\tok{darkblue}{ The}\tok{lightblue}{\textbackslash{}n}\tok{lightestblue}{ Hmm} cars are distinguishable,\tok{darkblue}{ meaning}\tok{lightblue}{ and} each\tok{darkblue}{ one}\tok{lightblue}{ car} is unique, and\tok{darkblue}{ the}\tok{lightblue}{ they}\tok{lightestblue}{ since} order in which they\tok{darkblue}{ approach}\tok{lightblue}{ arrive}\tok{darkblue}{ matters}\tok{lightblue}{ is} fixed\tok{darkblue}{.}\tok{lightblue}{.\textbackslash{}n\textbackslash{}n}\tok{darkblue}{ So}\tok{lightblue}{\textbackslash{}n}\tok{lightestblue}{ Hmm},\tok{darkblue}{ the}\tok{lightblue}{ car}\tok{lightestblue}{ different}\tok{darkblue}{ first}\tok{lightblue}{ cars} car\tok{darkblue}{ arrives}\tok{lightblue}{ gets}\tok{darkblue}{ first}\tok{lightblue}{,}, then the second, and so on\tok{darkblue}{ up}\tok{lightblue}{ until} to the sixth\tok{darkblue}{.}\tok{lightblue}{ car}\textbackslash{}n\tok{darkblue}{Hmm}\tok{lightblue}{First}.\tok{darkblue}{ Let}\tok{lightblue}{ So}\tok{darkblue}{ me}\tok{lightblue}{'s}\tok{darkblue}{ try}\tok{lightblue}{ break}\tok{lightestblue}{ start} to\tok{darkblue}{ break}\tok{lightblue}{ parse} this down.\tok{darkblue}{ Since}\tok{lightblue}{ First}\tok{darkblue}{ the}\tok{lightblue}{ each} car has three choices,%
\tok{darkblue}{ without}\tok{lightblue}{ if}\tok{lightestblue}{ I}\tok{darkblue}{ might}\tok{lightblue}{ could}\tok{lightestblue}{ guess} think\tok{darkblue}{ at}\tok{lightblue}{ initially} first\tok{darkblue}{ that}\tok{lightblue}{ it}\tok{darkblue}{ there}\tok{lightblue}{ the} total number of\tok{darkblue}{ possible}\tok{lightblue}{ ways}\tok{darkblue}{ ways}\tok{lightblue}{ arrangements} they can\tok{darkblue}{ choose}\tok{lightblue}{ stack}\tok{lightestblue}{ move}\tok{darkblue}{ their}\tok{lightblue}{ the} lanes is\tok{darkblue}{ }\tok{lightblue}{ just}3\textasciicircum{}6\tok{darkblue}{.}\tok{lightblue}{,}\tok{darkblue}{ But}\tok{lightblue}{ Let}\tok{lightestblue}{ Because} wait,\tok{darkblue}{ that}\tok{lightblue}{ the}\tok{darkblue}{ would}\tok{lightblue}{'s}\tok{lightestblue}{ includes}\tok{darkblue}{ be}\tok{lightblue}{ include} the case\tok{darkblue}{ without}\tok{lightblue}{ if}\tok{darkblue}{ the}\tok{lightblue}{ all}\tok{lightestblue}{ they}\tok{darkblue}{ order}\tok{lightblue}{ problem}\tok{darkblue}{ didn}\tok{lightblue}{ of}\tok{lightestblue}{ in}'t matter\tok{darkblue}{,}\tok{lightblue}{ or}\tok{darkblue}{ but}\tok{lightblue}{ right}\tok{darkblue}{ here}\tok{lightblue}{ since}\tok{darkblue}{,}\tok{lightblue}{ the}\tok{darkblue}{ since}\tok{lightblue}{ the}\tok{darkblue}{ order}\tok{lightblue}{ cars}\tok{darkblue}{ does}\tok{lightblue}{ in} matter because\tok{darkblue}{ the}\tok{lightblue}{ each} cars\tok{darkblue}{ are}\tok{lightblue}{ pull}\tok{darkblue}{ distinguish}\tok{lightblue}{ coming}able and\tok{darkblue}{ they}\tok{lightblue}{ arrive}\tok{lightestblue}{ pull}\tok{darkblue}{ arrive}\tok{lightblue}{ pull}\tok{lightestblue}{ come}\tok{darkblue}{ one}\tok{lightblue}{ sequentially}\tok{darkblue}{ after}\tok{lightblue}{ at} another.\tok{darkblue}{ So}\tok{lightblue}{ Hmm}\tok{lightestblue}{ However}\tok{darkblue}{,}\tok{lightblue}{ maybe}\tok{darkblue}{ maybe}\tok{lightblue}{ that}\tok{darkblue}{ each}\tok{lightblue}{ considering}\tok{darkblue}{ arrangement}\tok{lightblue}{ car} is\tok{darkblue}{ a}\tok{lightblue}{ just}\tok{darkblue}{ permutation}\tok{lightblue}{ sequence}\tok{darkblue}{ with}\tok{lightblue}{ where}\tok{darkblue}{ repetition}\tok{lightblue}{ specific}\tok{darkblue}{?}\tok{lightblue}{?\textbackslash{}n\textbackslash{}n}\tok{darkblue}{ So}\tok{lightblue}{\textbackslash{}n}\tok{darkblue}{,}\tok{lightblue}{ yes}\tok{darkblue}{ yeah}\tok{lightblue}{ thinking},\tok{darkblue}{ so}\tok{lightblue}{ }\tok{lightestblue}{ I}\tok{darkblue}{ each}\tok{lightblue}{ }\tok{lightestblue}{ that}\tok{darkblue}{ total}\tok{lightblue}{ each}\tok{lightestblue}{ three}\tok{darkblue}{ number}\tok{lightblue}{ without}\tok{darkblue}{ of}\tok{lightblue}{ is}\tok{lightestblue}{ should}\tok{darkblue}{ ways}\tok{lightblue}{ possible}\tok{lightestblue}{ possibilities}\tok{darkblue}{ without}\tok{lightblue}{ is}\tok{lightestblue}{ should}\tok{darkblue}{ any}\tok{lightblue}{ the} restrictions\tok{darkblue}{ would}\tok{lightblue}{ is}\tok{darkblue}{ be}\tok{lightblue}{ indeed} 3\tok{darkblue}{ multiplied}\tok{lightblue}{\textasciicircum{}}6\tok{darkblue}{,}\tok{lightblue}{.} which is 729\tok{darkblue}{.\textbackslash{}n\textbackslash{}n}\tok{lightblue}{.} But\tok{darkblue}{ the}\tok{lightblue}{ we}\tok{lightestblue}{ that} problem specifies that\tok{darkblue}{ all}\tok{lightblue}{ we} three lanes must be occupied.\tok{darkblue}{ That}\tok{lightblue}{ So}\tok{darkblue}{,}\tok{lightblue}{ I} I\tok{darkblue}{ need}\tok{lightblue}{ can} to subtract the\tok{darkblue}{ cases}\tok{lightblue}{ scenarios} where one or more%
 lanes are empty\tok{darkblue}{.\textbackslash{}n\textbackslash{}n}\tok{lightblue}{.}This\tok{darkblue}{ sounds}\tok{lightblue}{ seems}\tok{lightestblue}{ feels} like\tok{darkblue}{ an}\tok{lightblue}{ a} inclusion-exclusion\tok{darkblue}{ problem}\tok{lightblue}{ principle}.\tok{darkblue}{ In}\tok{lightblue}{ So}\tok{lightestblue}{ I}\tok{darkblue}{clusion}\tok{lightblue}{ comb}\tok{lightestblue}{ general}-exclusion is\tok{darkblue}{ a}\tok{lightblue}{ used}\tok{darkblue}{ principle}\tok{lightblue}{ counting}\tok{lightestblue}{ method}\tok{darkblue}{ where}\tok{lightblue}{ that}\tok{lightestblue}{ in}\tok{darkblue}{ you}\tok{lightblue}{ we}\tok{darkblue}{ can}\tok{lightblue}{ subtract}\tok{darkblue}{ calculate}\tok{lightblue}{ find} the\tok{darkblue}{ number}\tok{lightblue}{ total} of elements in a\tok{darkblue}{ union}\tok{lightblue}{ set}\tok{darkblue}{ of}\tok{lightblue}{ by}\tok{darkblue}{ overlapping}\tok{lightblue}{ sets} sets by\tok{darkblue}{ including}\tok{lightblue}{ adding}\tok{darkblue}{ and}\tok{lightblue}{ the} excluding the sizes of various intersections. So,\tok{darkblue}{ applying}\tok{lightblue}{ maybe}\tok{lightestblue}{ I}\tok{darkblue}{ that}\tok{lightblue}{ this}\tok{darkblue}{ here}\tok{lightblue}{ idea},\tok{darkblue}{ I}\tok{lightblue}{ perhaps}\tok{darkblue}{ need}\tok{lightblue}{ can} to subtract\tok{darkblue}{ the}\tok{lightblue}{ all}\tok{darkblue}{ cases}\tok{lightblue}{ number} where\tok{darkblue}{ one}\tok{lightblue}{ at}\tok{lightestblue}{ only} lane is empty,\tok{darkblue}{ but}\tok{lightblue}{ add} then add back in the cases where two lanes are empty\tok{darkblue}{ because}\tok{lightblue}{ since}\tok{darkblue}{ those}\tok{lightblue}{ we}\tok{lightestblue}{ I}\tok{darkblue}{ were}\tok{lightblue}{ have}\tok{lightestblue}{ cases} subtracted too many times.\textbackslash{}n\textbackslash{}n\tok{darkblue}{Let}\tok{lightblue}{Alternatively}\tok{darkblue}{'s}\tok{lightblue}{ me}\tok{darkblue}{ recall}\tok{lightblue}{ write}\tok{lightestblue}{ structure}\tok{darkblue}{ this}\tok{lightblue}{ that}\tok{darkblue}{ out}\tok{lightblue}{ formal}\tok{darkblue}{.\textbackslash{}n\textbackslash{}n}\tok{lightblue}{:\textbackslash{}n\textbackslash{}n}\tok{darkblue}{First}\tok{lightblue}{Total},\tok{darkblue}{ total}\tok{lightblue}{ the} number of ways\tok{darkblue}{ without}\tok{lightblue}{:}\tok{darkblue}{ any}\tok{lightblue}{ restrictions} restrictions\tok{darkblue}{:}\tok{lightblue}{ is} 3\textasciicircum{}6\tok{darkblue}{ =}\tok{lightblue}{.\textbackslash{}n\textbackslash{}n} 729.\textbackslash{}n\textbackslash{}nNow,\tok{darkblue}{ subtract}\tok{lightblue}{ the}\tok{lightestblue}{ compute} the\tok{darkblue}{ number}\tok{lightblue}{ cases} of ways where\tok{darkblue}{ at}\tok{lightblue}{ one} least one%
 lane is empty\tok{darkblue}{.\textbackslash{}n\textbackslash{}n}\tok{lightblue}{.} There are three lanes, so\tok{darkblue}{ the}\tok{lightblue}{ how} number of ways where\tok{darkblue}{ the}\tok{lightblue}{ a} left-turn lane is empty\tok{darkblue}{ is}\tok{lightblue}{,} 2\textasciicircum{}6\tok{darkblue}{,}\tok{lightblue}{ because}\tok{darkblue}{ since}\tok{lightblue}{ because} each car\tok{darkblue}{ can}\tok{lightblue}{ has}\tok{darkblue}{ only}\tok{lightblue}{ either}\tok{darkblue}{ choose}\tok{lightblue}{ go} between straight\tok{darkblue}{-going}\tok{lightblue}{ or} and right-turn lanes. Similarly, the\tok{darkblue}{ number}\tok{lightblue}{ same} of ways where the straight-going lane is empty is 2\textasciicircum{}6, and the number of ways where the right-turn lane is empty is\tok{darkblue}{ also}\tok{lightblue}{ } 2\textasciicircum{}6.\textbackslash{}n\textbackslash{}nSo,\tok{darkblue}{ subtract}\tok{lightblue}{ each}ing these\tok{darkblue}{,}\tok{lightblue}{:}\tok{darkblue}{ we}\tok{lightblue}{ it}\tok{darkblue}{ get}\tok{lightblue}{ have}\tok{darkblue}{ }\tok{lightblue}{:\textbackslash{}n\textbackslash{}n}\tok{darkblue}{3}\tok{lightblue}{7}*(2\textasciicircum{}6\tok{darkblue}{).}\tok{lightblue}{)} = 3*64 = 192\tok{darkblue}{.}\tok{lightblue}{.\textbackslash{}n\textbackslash{}n}But\tok{darkblue}{ wait}\tok{lightblue}{ now},\tok{darkblue}{ now}\tok{lightblue}{ inclusion}\tok{lightestblue}{ by} we've subtracted too much\tok{darkblue}{ because}\tok{lightblue}{.}\tok{lightestblue}{,}\tok{darkblue}{ the}\tok{lightblue}{ when} cases where%
 two lanes are empty have been subtracted\tok{darkblue}{ multiple}\tok{lightblue}{ three} times.\tok{darkblue}{ So}\tok{lightblue}{ For}\tok{lightestblue}{ Each}, we\tok{darkblue}{ need}\tok{lightblue}{ have} to add\tok{darkblue}{ them}\tok{lightblue}{ those} back in.\textbackslash{}n\textbackslash{}n\tok{darkblue}{The}\tok{lightblue}{How} many ways are there where two lanes are empty?\tok{darkblue}{ For}\tok{lightblue}{ Well}\tok{darkblue}{ each}\tok{lightblue}{ example}, if both the left-turn and straight-going lanes are empty,\tok{darkblue}{ all}\tok{lightblue}{ then}\tok{darkblue}{ cars}\tok{lightblue}{ six}\tok{darkblue}{ must}\tok{lightblue}{ would} go\tok{darkblue}{ right}\tok{lightblue}{ straight}.\tok{darkblue}{ There}\tok{lightblue}{ Similarly}'s only\tok{darkblue}{ }\tok{lightblue}{ one}1 way for\tok{darkblue}{ each}\tok{lightblue}{ that}\tok{darkblue}{ pair}\tok{lightblue}{ such}\tok{lightestblue}{ lane} of lanes\tok{darkblue}{.}\tok{lightblue}{;}\tok{darkblue}{ Since}\tok{lightblue}{ There} there are C(3,2) = 3\tok{darkblue}{ pairs}\tok{lightblue}{ such}\tok{lightestblue}{ ways} of lanes\tok{darkblue}{,}\tok{lightblue}{ (} each\tok{darkblue}{ contributing}\tok{lightblue}{ with} 1\tok{darkblue}{ way}\tok{lightblue}{\textasciicircum{}}\tok{darkblue}{,}\tok{lightblue}{ (}\tok{lightestblue}{ where}\tok{darkblue}{ so}\tok{lightblue}{ we}\tok{darkblue}{ adding}\tok{lightblue}{ we} add\tok{darkblue}{ back}\tok{lightblue}{ } 3\tok{darkblue}{*(}\tok{lightblue}{*}1\tok{darkblue}{ =}\tok{lightblue}{\textasciicircum{}} 3.\textbackslash{}n\textbackslash{}n\tok{darkblue}{So}\tok{lightblue}{Therefore},\tok{darkblue}{ applying}\tok{lightblue}{ putting} inclusion-exclusion, the total number of\tok{darkblue}{ ways}\tok{lightblue}{ valid} where all three lanes are occupied is:\textbackslash{}n\textbackslash{}n\tok{darkblue}{Total}\tok{lightblue}{3}%
 = 3\textasciicircum{}6 - 3*(2\textasciicircum{}6) + 3*(1\tok{darkblue}{\textasciicircum{}}\tok{lightblue}{)\textasciicircum{}}6\tok{darkblue}{)\textbackslash{}n\textbackslash{}n}\tok{lightblue}{)}\tok{lightestblue}{)\textbackslash{}n}\tok{darkblue}{Calcul}\tok{lightblue}{Pl}ating\tok{darkblue}{ that}\tok{lightblue}{:\textbackslash{}n\textbackslash{}n}:\textbackslash{}n\textbackslash{}n3\textasciicircum{}6 = 729\textbackslash{}n\tok{darkblue}{3}\tok{lightblue}{2}\tok{darkblue}{*(}\tok{lightblue}{*}2\textasciicircum{}6) = \tok{darkblue}{3}\tok{lightblue}{1}*64 = 192\textbackslash{}n3*(1\textasciicircum{}6) = 3*1 = 3\textbackslash{}nSo\tok{darkblue}{,\textbackslash{}n\textbackslash{}n}\tok{lightblue}{,} Total = 729 - 192 + 3\textbackslash{}n\tok{darkblue}{Let}\tok{lightblue}{Calcul}\tok{lightestblue}{Comput} me compute that\tok{darkblue}{ step}\tok{lightblue}{:\textbackslash{}n\textbackslash{}n} by step\tok{darkblue}{:\textbackslash{}n\textbackslash{}n}\tok{lightblue}{.\textbackslash{}n\textbackslash{}n}729 - 192\tok{darkblue}{ =}\tok{lightblue}{ is} 537\textbackslash{}n537 + 3 = 540\textbackslash{}nSo,%
\tok{darkblue}{ the}\tok{lightblue}{ is}\tok{lightestblue}{ there} total number\tok{darkblue}{ is}\tok{lightblue}{ of} ways\tok{darkblue}{ is}\tok{lightblue}{ the} 540\tok{darkblue}{.\textbackslash{}n\textbackslash{}n}\tok{lightblue}{?}\tok{darkblue}{Wait}\tok{lightblue}{Hmm},\tok{darkblue}{ but}\tok{lightblue}{ hold}\tok{lightestblue}{ let}\tok{darkblue}{ let}\tok{lightblue}{ hold} me\tok{darkblue}{ think}\tok{lightblue}{ make}\tok{lightestblue}{ just} again\tok{darkblue}{.}\tok{lightblue}{.\textbackslash{}n\textbackslash{}n} Is\tok{darkblue}{ this}\tok{lightblue}{ that}\tok{lightestblue}{ the}\tok{darkblue}{ the}\tok{lightblue}{ correct}\tok{lightestblue}{ actually}\tok{darkblue}{ correct}\tok{lightblue}{ right}\tok{lightestblue}{ same} approach\tok{darkblue}{?\textbackslash{}n\textbackslash{}n}\tok{lightblue}{?}\tok{darkblue}{Alternatively}\tok{lightblue}{Each}\tok{lightestblue}{In}, another way to\tok{darkblue}{ think}\tok{lightblue}{ approach} about\tok{darkblue}{ it}\tok{lightblue}{ this} is\tok{darkblue}{ using}\tok{lightblue}{ assigning}\tok{darkblue}{ the}\tok{lightblue}{ permutations} principle of\tok{darkblue}{ inclusion}\tok{lightblue}{ counting}-exclusion\tok{darkblue}{ for}\tok{lightblue}{ but}\tok{lightestblue}{ over}\tok{darkblue}{ sur}\tok{lightblue}{ counting}jective functions.\tok{darkblue}{ Since}\tok{lightblue}{ In}\tok{darkblue}{ we}\tok{lightblue}{ each}\tok{darkblue}{ want}\tok{lightblue}{ have}\tok{lightestblue}{'re}\tok{darkblue}{ all}\tok{lightblue}{ to} three lanes\tok{darkblue}{ to}\tok{lightblue}{ occupied} have at least one car,\tok{darkblue}{ this}\tok{lightblue}{ we}\tok{lightestblue}{ which} is equivalent to\tok{darkblue}{ counting}\tok{lightblue}{ finding}\tok{lightestblue}{ the} the number of onto functions from the set of six cars to the\tok{darkblue}{ three}\tok{lightblue}{ set} lanes\tok{darkblue}{.}\tok{lightblue}{.\textbackslash{}n\textbackslash{}n}\tok{darkblue}{ The}\tok{lightblue}{\textbackslash{}n} formula for\tok{darkblue}{ that}\tok{lightblue}{ this} is\tok{darkblue}{ indeed}\tok{lightblue}{:\textbackslash{}n\textbackslash{}n}\tok{darkblue}{ }\tok{lightblue}{ the}3! * S(6,3), where S(6,3) is the Stirling numbers of the second kind\tok{darkblue}{.}\tok{lightblue}{,}\tok{darkblue}{ which}\tok{lightblue}{ representing}\tok{lightestblue}{ counting} count the number of ways to%
 partition a set of\tok{darkblue}{ n}\tok{lightblue}{ six}\tok{lightestblue}{ } objects into k non-empty subsets\tok{darkblue}{.\textbackslash{}n\textbackslash{}n}\tok{lightblue}{.}\tok{darkblue}{But}\tok{lightblue}{Let}\tok{darkblue}{ wait}\tok{lightblue}{ S}\tok{lightestblue}{ considering},\tok{darkblue}{ in}\tok{lightblue}{ actually}\tok{lightestblue}{ the}\tok{darkblue}{ our}\tok{lightblue}{ this} case,\tok{darkblue}{ the}\tok{lightblue}{ each}\tok{lightestblue}{ is} cars are\tok{darkblue}{ distinguish}\tok{lightblue}{ ordered}\tok{lightestblue}{ labeled}able\tok{darkblue}{,}\tok{lightblue}{ by}\tok{darkblue}{ and}\tok{lightblue}{ so} the\tok{darkblue}{ order}\tok{lightblue}{ lanes} are distinguishable\tok{darkblue}{,}\tok{lightblue}{.}\tok{darkblue}{ so}\tok{lightblue}{ meaning}\tok{darkblue}{ the}\tok{lightblue}{ yes}\tok{lightestblue}{ S}\tok{darkblue}{ number}\tok{lightblue}{ formula}\tok{darkblue}{ of}\tok{lightblue}{ should} onto functions is\tok{darkblue}{ indeed}\tok{lightblue}{ }\tok{lightestblue}{ given} 3! * S(6,3).\tok{darkblue}{\textbackslash{}n}\tok{lightblue}{ So}\tok{lightestblue}{ Then}\tok{darkblue}{Let}\tok{lightblue}{I}\tok{lightestblue}{Alternatively} me\tok{darkblue}{ recall}\tok{lightblue}{ check}\tok{lightestblue}{ calculate}\tok{darkblue}{,}\tok{lightblue}{ that}\tok{darkblue}{ St}\tok{lightblue}{ the}\tok{darkblue}{ formula}\tok{lightblue}{ number}\tok{darkblue}{ for}\tok{lightblue}{ is} Stirling numbers\tok{darkblue}{ of}\tok{lightblue}{ is} the second kind\tok{darkblue}{ S}\tok{lightblue}{ is} S(n,k) = S(n-1,k-1) + k*S(n-1,k).\tok{darkblue}{ So}\tok{lightblue}{ But},\tok{darkblue}{ perhaps}\tok{lightblue}{ let}\tok{lightestblue}{ I}\tok{darkblue}{ maybe}\tok{lightblue}{ could}\tok{darkblue}{ I}\tok{lightblue}{ calculating}\tok{lightestblue}{ using} can compute S(6,3\tok{darkblue}{).\textbackslash{}n\textbackslash{}n}\tok{lightblue}{)?\textbackslash{}n\textbackslash{}n}\tok{darkblue}{But}\tok{lightblue}{Starting}\tok{darkblue}{ perhaps}\tok{lightblue}{ do}\tok{lightestblue}{ another}\tok{darkblue}{ I}\tok{lightblue}{ instead}\tok{lightestblue}{ we}\tok{darkblue}{ can}\tok{lightblue}{ should}\tok{lightestblue}{'m}\tok{darkblue}{ compute}\tok{lightblue}{ recall}\tok{darkblue}{ it}\tok{lightblue}{ }\tok{lightestblue}{ S}\tok{darkblue}{ directly}\tok{lightblue}{ another}\tok{lightestblue}{ as}\tok{darkblue}{.}\tok{lightblue}{ or}\tok{lightestblue}{ instead}\tok{darkblue}{ Alternatively}\tok{lightblue}{ But},\tok{darkblue}{ I}\tok{lightblue}{ could}\tok{darkblue}{ can}\tok{lightblue}{ remember}\tok{darkblue}{ use}\tok{lightblue}{ compute}\tok{darkblue}{ inclusion}\tok{lightblue}{ the}\tok{lightestblue}{ generating}-exclusion\tok{darkblue}{ formula}\tok{lightblue}{,}\tok{darkblue}{ as}\tok{lightblue}{.\textbackslash{}n\textbackslash{}n}\tok{lightestblue}{ too}\tok{darkblue}{ before}\tok{lightblue}{ above}\tok{lightestblue}{ done}\tok{darkblue}{.\textbackslash{}n\textbackslash{}n}\tok{lightblue}{,}\tok{darkblue}{Wait}\tok{lightblue}{Hold},%
\tok{darkblue}{ but}\tok{lightblue}{ actually}\tok{lightestblue}{ if}\tok{darkblue}{ hold}\tok{lightblue}{ our}\tok{lightestblue}{ haven} on\tok{darkblue}{,}\tok{lightblue}{.}\tok{lightestblue}{:}\tok{darkblue}{ let}\tok{lightblue}{ through}\tok{lightestblue}{ conflict}\tok{darkblue}{ me}\tok{lightblue}{'s}\tok{darkblue}{ make}\tok{lightblue}{ confirm} sure\tok{darkblue}{ the}\tok{lightblue}{ about}\tok{lightestblue}{,}\tok{darkblue}{ the}\tok{lightblue}{ our}\tok{darkblue}{ formula}\tok{lightblue}{ results}\tok{darkblue}{.\textbackslash{}n\textbackslash{}n}\tok{lightblue}{.}\tok{darkblue}{The}\tok{lightblue}{Yes}\tok{lightestblue}{Indeed}\tok{darkblue}{ number}\tok{lightblue}{ inclusion} of onto functions from a set\tok{darkblue}{ of}\tok{lightblue}{ A} size n to a set of size k is\tok{darkblue}{ indeed}\tok{lightblue}{ given}\tok{lightestblue}{ equal}\tok{darkblue}{ k}\tok{lightblue}{ given}\tok{lightestblue}{:\textbackslash{}n\textbackslash{}n}! * S(n,k).\tok{darkblue}{ So}\tok{lightblue}{\textbackslash{}n},\tok{darkblue}{ for}\tok{lightblue}{ if}\tok{lightestblue}{ the}\tok{darkblue}{ n}\tok{lightblue}{ our}\tok{lightestblue}{ S}=6\tok{darkblue}{ and}\tok{lightblue}{,} k=3\tok{darkblue}{,}\tok{lightblue}{.\textbackslash{}n\textbackslash{}n} it\tok{darkblue}{ would}\tok{lightblue}{ is} be \tok{darkblue}{3}\tok{lightblue}{6}! * S(6,3\tok{darkblue}{).\textbackslash{}n\textbackslash{}n}\tok{lightblue}{).}\tok{darkblue}{But}\tok{lightblue}{Let}\tok{darkblue}{ what}\tok{lightblue}{ }\tok{lightestblue}{ perhaps} is S(6,3\tok{darkblue}{)?\textbackslash{}n\textbackslash{}n}\tok{lightblue}{)?}\tok{darkblue}{I}\tok{lightblue}{The}\tok{lightestblue}{Calcul}\tok{darkblue}{ think}\tok{lightblue}{ recall}\tok{darkblue}{ S}\tok{lightblue}{ it}\tok{lightestblue}{ if}(6,3)\tok{darkblue}{ can}\tok{lightblue}{ is} be calculated\tok{darkblue}{ using}\tok{lightblue}{ as}\tok{darkblue}{ inclusion}\tok{lightblue}{ formula}-exclusion\tok{darkblue}{ as}\tok{lightblue}{,}\tok{darkblue}{ well}\tok{lightblue}{:\textbackslash{}n\textbackslash{}n}\tok{darkblue}{.\textbackslash{}n\textbackslash{}n}\tok{lightblue}{.}\tok{darkblue}{Alternatively}\tok{lightblue}{S}\tok{lightestblue}{Rec},\tok{darkblue}{ I}\tok{lightblue}{ maybe}\tok{lightestblue}{ we}\tok{darkblue}{ can}\tok{lightblue}{ remember}\tok{darkblue}{ recall}\tok{lightblue}{ think}\tok{lightestblue}{ refer} that\tok{darkblue}{ S}\tok{lightblue}{:\textbackslash{}n\textbackslash{}n}(n,k)\tok{darkblue}{ is}\tok{lightblue}{ =}\tok{darkblue}{ (}\tok{lightblue}{ S}\tok{lightestblue}{ sum}1/k\tok{darkblue}{!)}\tok{lightblue}{!} *\tok{darkblue}{ sum}\tok{lightblue}{ Sum}\_\{i=0 to k\} (-1)\textasciicircum{}i * C(k,i)%
 * (k - i)\textasciicircum{}n\tok{darkblue}{\}\textbackslash{}n\textbackslash{}n}\tok{lightblue}{.\textbackslash{}n\textbackslash{}n}\tok{darkblue}{So}\tok{lightblue}{Wait}\tok{lightestblue}{But}\tok{darkblue}{,}\tok{lightblue}{ for}\tok{darkblue}{ applying}\tok{lightblue}{ for} S(6,3\tok{darkblue}{):\textbackslash{}n\textbackslash{}n}\tok{lightblue}{),}S(6,3) = (1/3!)\tok{darkblue}{ *}\tok{lightblue}{ [} [\tok{darkblue}{C}\tok{lightblue}{3}(3,0\tok{darkblue}{)*}\tok{lightblue}{)*(}3\textasciicircum{}6 - C(3,1)*2\textasciicircum{}6 + C(3,2)*1\textasciicircum{}6 - C(3,3)*0\textasciicircum{}6]\textbackslash{}n\textbackslash{}n\tok{darkblue}{Which}\tok{lightblue}{Wait}\tok{lightestblue}{So}\tok{darkblue}{ is}\tok{lightblue}{ simpl}\tok{darkblue}{ (}\tok{lightblue}{ same}1/6\tok{darkblue}{)[}\tok{lightblue}{)*(}\tok{darkblue}{1}\tok{lightblue}{7}*729 - 3*64 + 3*1 - 1*0] = (1/6\tok{darkblue}{)(}\tok{lightblue}{)[}729 - 192 + 3\tok{darkblue}{ -}\tok{lightblue}{)} =\tok{darkblue}{ (}\tok{lightblue}{ same}1/6)(540)%
 = 90.\textbackslash{}n\textbackslash{}n\tok{darkblue}{So}\tok{lightblue}{Therefore}\tok{darkblue}{,}\tok{lightblue}{ S}\tok{darkblue}{ S}\tok{lightblue}{ then}(6,3)\tok{darkblue}{ =}\tok{lightblue}{ is} 90\tok{darkblue}{.}\tok{lightblue}{.\textbackslash{}n\textbackslash{}n}\tok{darkblue}{ Therefore}\tok{lightblue}{ Then}, the number of onto functions is 3! * 90 = 6 * 90 = 540\tok{darkblue}{.\textbackslash{}n\textbackslash{}n}\tok{lightblue}{.}Which\tok{darkblue}{ matches}\tok{lightblue}{ is}\tok{lightestblue}{ confirms}\tok{darkblue}{ our}\tok{lightblue}{ the}\tok{darkblue}{ result}\tok{lightblue}{ earlier}\tok{darkblue}{ from}\tok{lightblue}{ we}\tok{lightestblue}{ before}\tok{darkblue}{ inclusion}\tok{lightblue}{ earlier}\tok{lightestblue}{ above}-exclusion\tok{darkblue}{ earlier}\tok{lightblue}{ above}. So\tok{darkblue}{,}\tok{lightblue}{ that}\tok{darkblue}{ that}\tok{lightblue}{ this}\tok{darkblue}{'s}\tok{lightblue}{ seems}\tok{lightestblue}{ confirms}\tok{darkblue}{ reassuring}\tok{lightblue}{ consistent}\tok{lightestblue}{ comforting}.\textbackslash{}n\textbackslash{}n\tok{darkblue}{So}\tok{lightblue}{H}\tok{lightestblue}{Thus},\tok{darkblue}{ both}\tok{lightblue}{ it}\tok{darkblue}{ methods}\tok{lightblue}{ approaches}\tok{darkblue}{ give}\tok{lightblue}{ lead}\tok{darkblue}{ me}\tok{lightblue}{ the} 540\tok{darkblue}{.\textbackslash{}n\textbackslash{}n}\tok{lightblue}{,}\tok{darkblue}{Wait}\tok{lightblue}{Is}, but\tok{darkblue}{ let}\tok{lightblue}{ just}\tok{lightestblue}{ is} me\tok{darkblue}{ try}\tok{lightblue}{ just}\tok{darkblue}{ think}\tok{lightblue}{ double}\tok{darkblue}{ through}\tok{lightblue}{ of}\tok{darkblue}{ this}\tok{lightblue}{ an}\tok{lightestblue}{,}\tok{darkblue}{ another}\tok{lightblue}{ different}\tok{darkblue}{ way}\tok{lightblue}{ perspective}\tok{darkblue}{ to}\tok{lightblue}{.}\tok{darkblue}{ make}\tok{lightblue}{ be}\tok{lightestblue}{ see} sure\tok{darkblue}{.\textbackslash{}n\textbackslash{}n}\tok{lightblue}{.}\tok{darkblue}{Sup}\tok{lightblue}{Alternatively}pose\tok{darkblue}{ I}\tok{lightblue}{ we}\tok{lightestblue}{ that}\tok{darkblue}{ model}\tok{lightblue}{ think}\tok{lightestblue}{ assign}\tok{darkblue}{ each}\tok{lightblue}{ this}\tok{darkblue}{ arrangement}\tok{lightblue}{ assignment}\tok{darkblue}{ of}\tok{lightblue}{ as}\tok{darkblue}{ assigning}\tok{lightblue}{ a} each\tok{darkblue}{ of}\tok{lightblue}{ car} the six cars to one of\tok{darkblue}{ the}\tok{lightblue}{ three} three lanes,\tok{darkblue}{ with}\tok{lightblue}{ such}\tok{darkblue}{ the}\tok{lightblue}{ each}\tok{darkblue}{ condition}\tok{lightblue}{ restriction} that\tok{darkblue}{ each}\tok{lightblue}{ all} lane\tok{darkblue}{ has}\tok{lightblue}{ must}\tok{lightestblue}{ is} at least one car.%
\tok{darkblue}{ So}\tok{lightblue}{ The},\tok{darkblue}{ the}\tok{lightblue}{ it}\tok{lightestblue}{ indeed} number\tok{darkblue}{ of}\tok{lightblue}{ is}\tok{darkblue}{ such}\tok{lightblue}{ assignments} assignments is\tok{darkblue}{ equal}\tok{lightblue}{ the}\tok{lightestblue}{ } to the\tok{darkblue}{ number}\tok{lightblue}{ inclusion} of ways to\tok{darkblue}{ distribute}\tok{lightblue}{ assign}\tok{darkblue}{ six}\tok{lightblue}{ the} distinguishable objects into three distinguishable boxes\tok{darkblue}{,}\tok{lightblue}{ with} each box containing at least one object.\textbackslash{}n\textbackslash{}n\tok{darkblue}{Which}\tok{lightblue}{Yes}\tok{lightestblue}{And}\tok{darkblue}{ is}\tok{lightblue}{ we}\tok{darkblue}{ exactly}\tok{lightblue}{ again}\tok{darkblue}{ the}\tok{lightblue}{ }\tok{darkblue}{ definition}\tok{lightblue}{ formula}\tok{lightestblue}{ inclusion} of\tok{darkblue}{ onto}\tok{lightblue}{ the}\tok{lightestblue}{ sur} functions\tok{darkblue}{,}\tok{lightblue}{ between}\tok{darkblue}{ so}\tok{lightblue}{ as}\tok{lightestblue}{ or}\tok{darkblue}{ again}\tok{lightblue}{ yeah}\tok{lightestblue}{ same}\tok{darkblue}{,}\tok{lightblue}{ }\tok{darkblue}{ }\tok{lightblue}{ onto}\tok{darkblue}{5}\tok{lightblue}{3}! * S(6,3\tok{darkblue}{)}\tok{lightblue}{),} = 540\tok{darkblue}{.\textbackslash{}n\textbackslash{}n}\tok{lightblue}{,}Alternatively,\tok{darkblue}{ another}\tok{lightblue}{ I}\tok{lightestblue}{ let} can\tok{darkblue}{ think}\tok{lightblue}{ compute} of\tok{darkblue}{ it}\tok{lightblue}{ the}\tok{lightestblue}{ inclusion} as\tok{darkblue}{ first}\tok{lightblue}{:\textbackslash{}n\textbackslash{}n}\tok{lightestblue}{ follows}\tok{darkblue}{ assigning}\tok{lightblue}{ choosing}\tok{lightestblue}{ ensuring} each of the six cars to\tok{darkblue}{ a}\tok{lightblue}{ one}\tok{lightestblue}{ any} lane\tok{darkblue}{,}\tok{lightblue}{ without}\tok{darkblue}{ then}\tok{lightblue}{ with} subtracting the\tok{darkblue}{ assignments}\tok{lightblue}{ cases}\tok{lightestblue}{ distributions} where\tok{darkblue}{ one}\tok{lightblue}{ a}\tok{lightestblue}{ some} or more lanes are empty.\tok{darkblue}{ Which}\tok{lightblue}{ But}\tok{darkblue}{ is}\tok{lightblue}{ leads}\tok{lightestblue}{ still}\tok{darkblue}{ what}\tok{lightblue}{ precisely}\tok{darkblue}{ I}\tok{lightblue}{ we} did\tok{darkblue}{ initially}\tok{lightblue}{ earlier}\tok{darkblue}{,}\tok{lightblue}{.\textbackslash{}n\textbackslash{}n}\tok{darkblue}{ leading}\tok{lightblue}{ giving}\tok{lightestblue}{ with}\tok{darkblue}{ me}\tok{lightblue}{ to}\tok{darkblue}{ }\tok{lightblue}{ the}\tok{darkblue}{5}\tok{lightblue}{3}40\tok{darkblue}{.\textbackslash{}n\textbackslash{}n}\tok{lightblue}{ as}\tok{darkblue}{So}\tok{lightblue}{But}\tok{lightestblue}{Just},\tok{darkblue}{ seems}\tok{lightblue}{ perhaps}\tok{lightestblue}{ does}\tok{darkblue}{ like}\tok{lightblue}{ consistent}\tok{darkblue}{ the}\tok{lightblue}{ all}\tok{lightestblue}{ multiple} answer is\tok{darkblue}{ }\tok{lightblue}{ consistent}540.\textbackslash{}n\textbackslash{}n\tok{darkblue}{Wait}\tok{lightblue}{Alternatively},%
\tok{darkblue}{ but}\tok{lightblue}{ maybe}\tok{lightestblue}{ now}\tok{darkblue}{ just}\tok{lightblue}{ another}\tok{lightestblue}{ hang} to\tok{darkblue}{ double}\tok{lightblue}{ think}\tok{lightestblue}{ cross}-check,\tok{darkblue}{ let}\tok{lightblue}{ maybe}\tok{lightestblue}{ another} me\tok{darkblue}{ think}\tok{lightblue}{ consider}\tok{lightestblue}{ plug}\tok{darkblue}{ of}\tok{lightblue}{ if}\tok{lightestblue}{ in}\tok{darkblue}{ another}\tok{lightblue}{ a}\tok{lightestblue}{ arranging}\tok{darkblue}{ smaller}\tok{lightblue}{ small}\tok{lightestblue}{ more} case\tok{darkblue}{.\textbackslash{}n\textbackslash{}n}\tok{lightblue}{ where}\tok{lightestblue}{,}Suppose\tok{darkblue}{ instead}\tok{lightblue}{ there} of\tok{darkblue}{ six}\tok{lightblue}{ } cars,\tok{darkblue}{ we}\tok{lightblue}{ let}\tok{lightestblue}{ say} have two cars\tok{darkblue}{ and}\tok{lightblue}{.\textbackslash{}n\textbackslash{}n}\tok{darkblue}{ three}\tok{lightblue}{ two} lanes\tok{darkblue}{.}\tok{lightblue}{,}\tok{darkblue}{ How}\tok{lightblue}{ Then}\tok{lightestblue}{ By} many ways\tok{darkblue}{ can}\tok{lightblue}{ are}\tok{darkblue}{ the}\tok{lightblue}{ they} stack up\tok{darkblue}{ so}\tok{lightblue}{ with}\tok{lightestblue}{ such}\tok{darkblue}{ both}\tok{lightblue}{ that} both lanes are occupied\tok{darkblue}{.\textbackslash{}n\textbackslash{}n}\tok{lightblue}{?\textbackslash{}n\textbackslash{}n}\tok{darkblue}{By}\tok{lightblue}{Using}\tok{darkblue}{ the}\tok{lightblue}{ our}\tok{darkblue}{ same}\tok{lightblue}{ formula}\tok{darkblue}{ logic}\tok{lightblue}{ method}\tok{lightestblue}{ reasoning}\tok{darkblue}{:}\tok{lightblue}{:\textbackslash{}n\textbackslash{}n}\tok{darkblue}{ total}\tok{lightblue}{ }\tok{darkblue}{ ways}\tok{lightblue}{ assignments}\tok{lightestblue}{ number}\tok{darkblue}{ without}\tok{lightblue}{ =}\tok{darkblue}{ restriction}\tok{lightblue}{ restrictions}\tok{darkblue}{:}\tok{lightblue}{ is} 2\textasciicircum{}2 = 4\tok{darkblue}{.\textbackslash{}n\textbackslash{}n}\tok{lightblue}{.}\tok{darkblue}{Sub}\tok{lightblue}{Number}tract the\tok{darkblue}{ two}\tok{lightblue}{ cases} where one lane is empty: 2\tok{darkblue}{*(}\tok{lightblue}{ *}1\textasciicircum{}2) = 2.\textbackslash{}n\textbackslash{}nSo,\tok{darkblue}{ total}\tok{lightblue}{ }\tok{lightestblue}{ inclusion}\tok{darkblue}{ =}\tok{lightblue}{ ways} 4 - 2 = 2\tok{darkblue}{.\textbackslash{}n\textbackslash{}n}\tok{lightblue}{.}\tok{darkblue}{Which}\tok{lightblue}{But}\tok{darkblue}{ is}\tok{lightblue}{ makes}\tok{lightestblue}{ indeed} correct\tok{darkblue}{,}\tok{lightblue}{:}\tok{darkblue}{ because}\tok{lightblue}{ since}\tok{darkblue}{ the}\tok{lightblue}{ each} two\tok{darkblue}{ cars}\tok{lightblue}{ possible}\tok{darkblue}{ can}\tok{lightblue}{ must}\tok{lightestblue}{ have}\tok{darkblue}{ be}\tok{lightblue}{ either}\tok{darkblue}{ in}\tok{lightblue}{ split}\tok{darkblue}{ two}\tok{lightblue}{ either} different\tok{darkblue}{ orders}\tok{lightblue}{ lanes}\tok{darkblue}{:}\tok{lightblue}{ (}\tok{lightestblue}{ with}\tok{darkblue}{ first}\tok{lightblue}{ one} car left, second\tok{darkblue}{ straight}\tok{lightblue}{ right}\tok{darkblue}{;}\tok{lightblue}{ or} or first car%
 straight, second left\tok{darkblue}{.}\tok{lightblue}{.\textbackslash{}n\textbackslash{}n}Alternatively,\tok{darkblue}{ with}\tok{lightblue}{ the}\tok{lightestblue}{ another}\tok{darkblue}{ number}\tok{lightblue}{ onto} of onto functions\tok{darkblue}{ from}\tok{lightblue}{ is} 2 cars to 2 lanes is 2! * S(2,2) = 2\tok{darkblue}{ *}\tok{lightblue}{*} 1 = 2\tok{darkblue}{,}\tok{lightblue}{.}\tok{lightestblue}{.\textbackslash{}n\textbackslash{}n} which\tok{darkblue}{ is}\tok{lightblue}{ matches}\tok{darkblue}{ correct}\tok{lightblue}{ consistent}.\textbackslash{}n\textbackslash{}n\tok{darkblue}{Another}\tok{lightblue}{Similarly},\tok{darkblue}{ let}\tok{lightblue}{ another}\tok{lightestblue}{ consider}\tok{darkblue}{ me}\tok{lightblue}{ us}\tok{darkblue}{ test}\tok{lightblue}{ think}\tok{darkblue}{ with}\tok{lightblue}{ three}\tok{lightestblue}{ for}\tok{darkblue}{ three}\tok{lightblue}{ n} cars and\tok{darkblue}{ two}\tok{lightblue}{ three} lanes\tok{darkblue}{.\textbackslash{}n\textbackslash{}n}\tok{lightblue}{.}Total\tok{darkblue}{ ways}\tok{lightblue}{ without}: 2\textasciicircum{}3\tok{darkblue}{ =}\tok{lightblue}{=} 8.\textbackslash{}n\textbackslash{}n\tok{darkblue}{Sub}\tok{lightblue}{Number}tract cases\tok{darkblue}{ where}\tok{lightblue}{ with} one lane is empty: 2*(1\textasciicircum{}3) = 2.\textbackslash{}n\textbackslash{}n\tok{darkblue}{So}\tok{lightblue}{Total}\tok{lightestblue}{Thus},\tok{darkblue}{ total}\tok{lightblue}{ }\tok{lightestblue}{ number}\tok{darkblue}{ onto}\tok{lightblue}{ =} 8 - 2 = 6.\textbackslash{}n\textbackslash{}nBut\tok{darkblue}{ actually}\tok{lightblue}{ how}\tok{lightestblue}{ also},\tok{darkblue}{ the}\tok{lightblue}{ using} number of\tok{darkblue}{ sur}\tok{lightblue}{ ways}\tok{darkblue}{ where}\tok{lightblue}{ is} both lanes are occupied\tok{darkblue}{ is}\tok{lightblue}{ with}\tok{darkblue}{ }\tok{lightblue}{ the}\tok{darkblue}{6}\tok{lightblue}{2}\tok{darkblue}{,}\tok{lightblue}{.}\tok{darkblue}{ which}\tok{lightblue}{ right}\tok{darkblue}{ is}\tok{lightblue}{ makes}\tok{lightestblue}{ matches}\tok{darkblue}{ }\tok{lightblue}{ also}\tok{lightestblue}{ equivalent}2!%
 * S(3,2) = 2 * 3 = 6\tok{darkblue}{.}\tok{lightblue}{,}\tok{darkblue}{ Which}\tok{lightblue}{ That}\tok{lightestblue}{ Also} is correct\tok{darkblue}{.\textbackslash{}n\textbackslash{}n}\tok{lightblue}{.}\tok{lightestblue}{ since}\tok{darkblue}{So}\tok{lightblue}{For},\tok{darkblue}{ the}\tok{lightblue}{ applying}\tok{darkblue}{ inclusion}\tok{lightblue}{ formula}\tok{lightestblue}{ method}\tok{darkblue}{ seems}\tok{lightblue}{ works}\tok{darkblue}{ to}\tok{lightblue}{ consistent} hold\tok{darkblue}{ up}\tok{lightblue}{.\textbackslash{}n\textbackslash{}n}\tok{darkblue}{Similarly}\tok{lightblue}{So}\tok{lightestblue}{Wait},\tok{darkblue}{ for}\tok{lightblue}{ in}\tok{lightestblue}{ if} the\tok{darkblue}{ case}\tok{lightblue}{ problem}\tok{darkblue}{ of}\tok{lightblue}{ where}\tok{darkblue}{ three}\tok{lightblue}{ our} cars and three lanes\tok{darkblue}{,}\tok{lightblue}{.\textbackslash{}n\textbackslash{}n}\tok{darkblue}{ the}\tok{lightblue}{ number}\tok{darkblue}{ number}\tok{lightblue}{ total} of\tok{darkblue}{ onto}\tok{lightblue}{ ways}\tok{lightestblue}{ assignments} functions\tok{darkblue}{ would}\tok{lightblue}{ is} 3! * S(3,3) = 6 * 1 = 6.\textbackslash{}n\textbackslash{}nWhich\tok{darkblue}{ is}\tok{lightblue}{ makes}\tok{lightestblue}{ also}\tok{darkblue}{ correct}\tok{lightblue}{ the}\tok{darkblue}{,}\tok{lightblue}{ since}\tok{lightestblue}{:}\tok{darkblue}{ since}\tok{lightblue}{ because}\tok{lightestblue}{ as} each\tok{darkblue}{ car}\tok{lightblue}{ lane}\tok{darkblue}{ can}\tok{lightblue}{ must} be assigned\tok{darkblue}{ uniquely}\tok{lightblue}{ to} to\tok{darkblue}{ a}\tok{lightblue}{ each} lane\tok{darkblue}{,}\tok{lightblue}{.\textbackslash{}n\textbackslash{}n}\tok{darkblue}{ so}\tok{lightblue}{ and}\tok{darkblue}{ }\tok{lightblue}{ permutations}3!\tok{darkblue}{ =}\tok{lightblue}{ ways} 6\tok{darkblue}{.\textbackslash{}n\textbackslash{}n}\tok{lightblue}{,}\tok{darkblue}{Therefore}\tok{lightblue}{Alternatively}\tok{lightestblue}{H},\tok{darkblue}{ the}\tok{lightblue}{ going}\tok{lightestblue}{ coming}\tok{darkblue}{ same}\tok{lightblue}{ formula}\tok{darkblue}{ seems}\tok{lightblue}{ is}\tok{lightestblue}{ holds}\tok{darkblue}{ to}\tok{lightblue}{ consistent}\tok{lightestblue}{ solid}\tok{darkblue}{.\textbackslash{}n\textbackslash{}n}\tok{lightblue}{ for}\tok{darkblue}{Therefore}\tok{lightblue}{Thus},\tok{darkblue}{ going}\tok{lightblue}{ applying}\tok{lightestblue}{ when} back,\tok{darkblue}{ for}\tok{lightblue}{ with}\tok{darkblue}{ six}\tok{lightblue}{ } cars and three lanes, the number\tok{darkblue}{ is}\tok{lightblue}{ of} \tok{darkblue}{5}\tok{lightblue}{3}40.\textbackslash{}n\textbackslash{}n\tok{darkblue}{Therefore}\tok{lightblue}{But}, I%
\tok{darkblue}{ think}\tok{lightblue}{ can}\tok{darkblue}{ I}\tok{lightblue}{ the}\tok{darkblue}{ can}\tok{lightblue}{'m}\tok{darkblue}{ confidently}\tok{lightblue}{ be} confident\tok{darkblue}{ that}\tok{lightblue}{ the}\tok{darkblue}{ the}\tok{lightblue}{ } answer is 540.\textbackslash{}n\textbackslash{}n\tok{darkblue}{Wait}\tok{lightblue}{Just}, but\tok{darkblue}{ just}\tok{lightblue}{ let}\tok{darkblue}{ to}\tok{lightblue}{ another}\tok{darkblue}{ make}\tok{lightblue}{ challenge}\tok{darkblue}{ sure}\tok{lightblue}{ entirely}\tok{darkblue}{,}\tok{lightblue}{ I}\tok{darkblue}{ perhaps}\tok{lightblue}{ maybe}\tok{lightestblue}{ I} I\tok{darkblue}{ can}\tok{lightblue}{ should}\tok{lightestblue}{ mis}\tok{darkblue}{ think}\tok{lightblue}{ compute}\tok{darkblue}{ of}\tok{lightblue}{ about}\tok{lightestblue}{ in}\tok{darkblue}{ it}\tok{lightblue}{ another}\tok{lightestblue}{ arranging}\tok{darkblue}{ as}\tok{lightblue}{ in}\tok{darkblue}{ assigning}\tok{lightblue}{ arranging}\tok{darkblue}{ each}\tok{lightblue}{ the}\tok{darkblue}{ car}\tok{lightblue}{ position}\tok{darkblue}{ to}\tok{lightblue}{ independently} a lane,\tok{darkblue}{ and}\tok{lightblue}{ then}\tok{lightestblue}{ which}\tok{darkblue}{ then}\tok{lightblue}{ using}\tok{lightestblue}{ ordering}\tok{darkblue}{ subtract}\tok{lightblue}{ computing}\tok{lightestblue}{ accounting}ing\tok{darkblue}{ the}\tok{lightblue}{ those} cases where\tok{darkblue}{ one}\tok{lightblue}{ lanes}\tok{darkblue}{ or}\tok{lightblue}{ lane} more lanes are empty.\textbackslash{}n\textbackslash{}n\tok{darkblue}{So}\tok{lightblue}{Total},\tok{darkblue}{ similar}\tok{lightblue}{ same}\tok{lightestblue}{ there}\tok{darkblue}{ as}\tok{lightblue}{ steps}\tok{darkblue}{ inclusion}\tok{lightblue}{ In}-exclusion\tok{darkblue}{.\textbackslash{}n\textbackslash{}n}\tok{lightblue}{:\textbackslash{}n\textbackslash{}n}\tok{lightestblue}{.}\tok{darkblue}{Total}\tok{lightblue}{Number} number of\tok{darkblue}{ assignments}\tok{lightblue}{ functions}\tok{lightestblue}{ ways}: 3\textasciicircum{}6\tok{darkblue}{ =}\tok{lightblue}{.\textbackslash{}n\textbackslash{}n} 729.\textbackslash{}n\textbackslash{}nNumber of assignments\tok{darkblue}{ where}\tok{lightblue}{ with}\tok{darkblue}{ left}\tok{lightblue}{ the}\tok{lightestblue}{ a}\tok{darkblue}{ lane}\tok{lightblue}{-turn}\tok{darkblue}{ is}\tok{lightblue}{ lane} empty: 2\textasciicircum{}6 = 64.\textbackslash{}n\textbackslash{}n\tok{darkblue}{Similarly}\tok{lightblue}{Number},\tok{darkblue}{ assignments}\tok{lightblue}{ where} where straight-going is empty: 64\tok{darkblue}{.\textbackslash{}n\textbackslash{}n}\tok{lightblue}{,}\tok{darkblue}{And}\tok{lightblue}{Assign}\tok{lightestblue}{Similarly}\tok{darkblue}{ assignments}\tok{lightblue}{,} where right-turn is empty: 64.\textbackslash{}n\textbackslash{}n\tok{darkblue}{So}\tok{lightblue}{Thus}, subtract\tok{darkblue}{ }\tok{lightblue}{ these}\tok{lightestblue}{ those}3*64%
 = 192\tok{darkblue}{.\textbackslash{}n\textbackslash{}n}\tok{lightblue}{,}But\tok{darkblue}{ now}\tok{lightblue}{ wait}\tok{lightestblue}{ as},\tok{darkblue}{ we}\tok{lightblue}{ when}\tok{lightestblue}{ assignments}\tok{darkblue}{ have}\tok{lightblue}{'ve} subtracted too much\tok{darkblue}{ because}\tok{lightblue}{ those} the\tok{darkblue}{ cases}\tok{lightblue}{ assignments} where two lanes are empty\tok{darkblue}{ have}\tok{lightblue}{ were} been subtracted\tok{darkblue}{ twice}\tok{lightblue}{ multiple} times\tok{darkblue}{.\textbackslash{}n\textbackslash{}n}\tok{lightblue}{.}\tok{darkblue}{So}\tok{lightblue}{Number},\tok{darkblue}{ how}\tok{lightblue}{ we} many\tok{darkblue}{ assignments}\tok{lightblue}{ cases} have\tok{darkblue}{ two}\tok{lightblue}{ both} lanes empty?\textbackslash{}n\textbackslash{}n\tok{darkblue}{If}\tok{lightblue}{For}\tok{darkblue}{ left}\tok{lightblue}{ two} and straight are empty\tok{darkblue}{:}\tok{lightblue}{,} all\tok{darkblue}{ six}\tok{lightblue}{ cars}\tok{darkblue}{ must}\tok{lightblue}{ go} go right:\tok{darkblue}{ }\tok{lightblue}{ only}1\tok{darkblue}{ way}\tok{lightblue}{ assignment}.\textbackslash{}n\textbackslash{}nSimilarly,\tok{darkblue}{ left}\tok{lightblue}{ if} and right\tok{darkblue}{ empty}\tok{lightblue}{ are}: all cars\tok{darkblue}{ go}\tok{lightblue}{ straight}\tok{lightestblue}{ must} straight\tok{darkblue}{:}\tok{lightblue}{.\textbackslash{}n\textbackslash{}n} 1 way.\textbackslash{}n\textbackslash{}nStraight and right empty: all cars go left: 1 way.\textbackslash{}n\textbackslash{}n\tok{darkblue}{So}\tok{lightblue}{Thus}\tok{lightestblue}{Total},\tok{darkblue}{ three}\tok{lightblue}{ for}\tok{darkblue}{ such}\tok{lightblue}{ assignments}\tok{darkblue}{ cases}\tok{lightblue}{ assignments}, each\tok{darkblue}{ contributing}\tok{lightblue}{ with} 1\tok{darkblue}{.\textbackslash{}n\textbackslash{}n}\tok{lightblue}{.}\tok{darkblue}{So}\tok{lightblue}{Therefore}, we\tok{darkblue}{ have}\tok{lightblue}{ add}\tok{darkblue}{ to}\tok{lightblue}{ subtract} add\tok{darkblue}{ them}\tok{lightblue}{ those} back\tok{darkblue}{.}\tok{lightblue}{ in}\tok{darkblue}{.}\tok{lightblue}{.\textbackslash{}n\textbackslash{}n}\tok{lightestblue}{,} So,\tok{darkblue}{ add}\tok{lightblue}{ +} 3*1 = 3.\textbackslash{}n\textbackslash{}n\tok{darkblue}{Therefore}\tok{lightblue}{And},%
\tok{darkblue}{ total}\tok{lightblue}{ so}\tok{darkblue}{ number}\tok{lightblue}{ is}\tok{lightestblue}{ =}\tok{darkblue}{ of}\tok{lightblue}{ is}\tok{darkblue}{ assignments}\tok{lightblue}{ onto}\tok{darkblue}{ where}\tok{lightblue}{ with}\tok{lightestblue}{ is} no\tok{darkblue}{ lanes}\tok{lightblue}{ lane} are empty is\tok{darkblue}{ }\tok{lightblue}{:\textbackslash{}n\textbackslash{}n}729 - 192 + 3 = 540.\textbackslash{}n\textbackslash{}n\tok{darkblue}{Yes}\tok{lightblue}{Yep}\tok{lightestblue}{Therefore},\tok{darkblue}{ so}\tok{lightblue}{ that}\tok{darkblue}{ matches}\tok{lightblue}{ is}\tok{darkblue}{ again}\tok{lightblue}{ our}\tok{lightestblue}{ as}.\textbackslash{}n\textbackslash{}n\tok{darkblue}{Alternatively}\tok{lightblue}{Therefore}\tok{lightestblue}{So},\tok{darkblue}{ I}\tok{lightblue}{ seems}\tok{lightestblue}{ regardless}\tok{darkblue}{ can}\tok{lightblue}{'m}\tok{lightestblue}{ don}\tok{darkblue}{ be}\tok{lightblue}{ confirm}\tok{darkblue}{ confident}\tok{lightblue}{ pretty}\tok{darkblue}{ that}\tok{lightblue}{ the}\tok{darkblue}{ }\tok{lightblue}{ the}540 is the correct answer.\textbackslash{}n\textbackslash{}n\tok{darkblue}{So}\tok{lightblue}{Thus}\tok{lightestblue}{Therefore},\tok{darkblue}{ the}\tok{lightblue}{ in}\tok{darkblue}{ number}\tok{lightblue}{ six}\tok{lightestblue}{ way} of ways\tok{darkblue}{ the}\tok{lightblue}{ is}\tok{darkblue}{ six}\tok{lightblue}{ cars} cars can stack up\tok{darkblue}{ at}\tok{lightblue}{ so} the intersection\tok{darkblue}{ with}\tok{lightblue}{ such}\tok{lightestblue}{ so} all three lanes occupied is 540.\textbackslash{}n\textbackslash{}n**Final Answer**\textbackslash{}nThe number of ways\tok{darkblue}{ is}\tok{lightblue}{ the} \textbackslash{}boxed\{540\}.\textbackslash{}n</think>\textbackslash{}nTo determine the number of ways six distinguishable cars can stack up at a red light such that all three lanes (left-turn,%
 straight-going, and right-turn) are occupied, we use the principle of inclusion-exclusion.\textbackslash{}n\textbackslash{}n1. Calculate the total number of unrestricted ways the cars can choose their lanes:\textbackslash{}n  \textbackslash{}[\textbackslash{}n  3\textasciicircum{}6 = 729\textbackslash{}n  \textbackslash{}]\textbackslash{}n\textbackslash{}n2. Subtract the number of ways where at least one lane is empty. There are three lanes, so the number of ways for each lane to be empty is:\textbackslash{}n  \textbackslash{}[\textbackslash{}n  2\textasciicircum{}6 = 64\textbackslash{}n  \textbackslash{}]\textbackslash{}n  Therefore, total ways for one lane being empty:\textbackslash{}n  \textbackslash{}[\textbackslash{}n  3 \textbackslash{}times%
 64 = 192\textbackslash{}n  \textbackslash{}]\textbackslash{}n\textbackslash{}n3. Add back the cases where two lanes are empty, since these were subtracted too many times. There are three pairs of lanes, and each pair contributes 1 way:\textbackslash{}n  \textbackslash{}[\textbackslash{}n  3 \textbackslash{}times 1 = 3\textbackslash{}n  \textbackslash{}]\textbackslash{}n\textbackslash{}n4. Apply the inclusion-exclusion principle:\textbackslash{}n  \textbackslash{}[\textbackslash{}n  729 - 192 + 3 = 540\textbackslash{}n  \textbackslash{}]\textbackslash{}n\textbackslash{}nThus, the number of ways the six cars can stack up so that all three lanes are occupied is \textbackslash{}(\textbackslash{}boxed\{540\}\textbackslash{}).<|end\_of\_sentence|>' (SPECIAL)

\end{tokenforkbox}

\end{document}